# LLM-coupled interpretable tool wear monitoring with multi-evidence reasoning in CNC machining

Yuntong Chen[1], Yingqi Li[1], Qiwen Shen[1], Guobin Zhao[2], Xitian Tian[1], Lijiang Huang[1]*
1 School of Mechanical Engineering, Northwestern Polytechnical University, Xi'an, Shaanxi, 710072, China
2 Department of Chemistry, National University of Singapore, 3 Science Drive 3, Singapore 117543, Singapore

E-mail addresses: chenyt123@mail.nwpu.edu.cn (Yuntong Chen), yingqili@mail.nwpu.edu.cn (Yingqi Li), shenqiwen@mail.nwpu.edu.cn(Qiwen Shen), guobinzhao@nus.edu.sg(Guobin Zhao), tianxt@nwpu.edu.cn (Xitian Tian), huanglj@nwpu.edu.cn (Lijiang Huang*).

* Corresponding author: Prof. Lijiang Huang* (huanglj@nwpu.edu.cn)

## Abstract

Interpretable tool wear assessment in CNC machining remains challenging because existing diagnostic models, despite achieving high classification accuracy, produce numerical outputs that may offer limited actionable guidance for practical replacement decisions. This study proposes a two-layer framework that aims to couple deep learning-based wear classification with LLM-assisted reasoning to support interpretable decision-making. The first layer processes multi-domain signal features through PCA-Loading-based feature selection and GRU classification with IPSO-SA optimization to identify three progressive wear states. Evaluation on the PHM2010 dataset identifies a Pareto-optimal configuration reaching 98.38% macro F1-score with 10 of 84 features within the tool wear trajectories, with performance insensitive to the choice of hyperparameter optimization method. The second layer employs an LLM-based module that receives pre-computed classification results, feature attributions, and temporal trend indicators through a structured interface, organizing them into natural language reports with replacement recommendations while confining safety-critical content to pre-computed fields to help ensure factual reliability. Evaluation across three LLMs shows consistent factual accuracy under this design. Blind evaluation by experienced CNC engineers rated LLM reports above both a basic and an enhanced rule template, with the enhanced template recovering most of the improvement, suggesting that organizing evidence into structured reports drives most of the decision-support value. Cross-dataset validation on 146 NASA Ames milling records with different workpiece materials, sensor configurations, and cutting conditions provides preliminary evidence that the framework can generalize without modification. These results suggest that coupling deep learning diagnostics with LLM-assisted reasoning offers a promising direction for enhancing interpretable decision support in tool wear management.

## 1. Introduction

Tool wear monitoring is a fundamental task in CNC machining. The wear state of cutting tools directly affects machining accuracy, surface quality, and production efficiency. Undetected wear progression can lead to workpiece scrap, unplanned downtime, and potential safety hazards. Consequently, accurate and timely assessment of tool condition has long been recognized as essential for cost reduction and quality assurance in modern manufacturing [1–3].

In recent years, data-driven approaches leveraging vibration signals and deep learning have become a dominant paradigm for tool wear assessment. Convolutional neural networks, recurrent architectures such as LSTM and GRU, and more recently Transformer-based models have been applied to extract temporal and spectral patterns from sensor data, achieving classification accuracies above 95% on benchmark datasets [4–7]. Multi-domain feature extraction combining time-domain statistics, frequency-domain descriptors, and wavelet-based time-frequency representations has been shown to be effective for capturing the diverse signatures of progressive tool degradation [8,9]. Metaheuristic algorithms including particle swarm optimization and genetic algorithms have further been used to optimize model hyperparameters, reducing the reliance on manual configuration [10,11]. Despite these advances in diagnostic accuracy, a notable gap remains between model output and operational decision-making. Deep learning classifiers produce numerical predictions, probability vectors, or abstract feature representations that are meaningful to model developers but may not directly support the practical decision of whether and when to replace a tool.

This interpretability gap has motivated growing interest in explainable artificial intelligence methods for fault diagnosis and condition monitoring. Post-hoc attribution techniques such as SHAP, LIME, and attention visualization have been applied to quantify the contribution of individual input features to model predictions [12–14]. These methods provide valuable insight into model behavior and have been shown to improve trust among model developers. However, their outputs take the form of numerical attribution scores, bar charts, or heatmaps that may require considerable analytical expertise to interpret. At the CNC workstation level, a SHAP summary plot may not readily answer the practical question of whether and why a tool should be replaced. The gap lies not in the availability of attribution information but in the ability to synthesize it into actionable guidance for practical replacement decisions [15].

Large language models offer a promising mechanism for addressing this challenge. Their capacity for natural language reasoning and synthesis can potentially enable the integration of multiple diagnostic evidence sources into coherent, practically interpretable outputs that support replacement decision-making. Recent work has begun to explore LLM applications in fault diagnosis and prognostics, ranging from direct diagnostic classification using fine-tuned language models [16–18] to multimodal frameworks that align sensor signals with textual semantics [19,20]. Comprehensive surveys have mapped the emerging landscape of foundation models in PHM [21–23]. Physics-constrained and knowledge-grounded approaches have been proposed to improve diagnostic reliability [24,25], and trustworthy multi-agent LLM architectures have begun to address auditability and reasoning provenance [26]. However, existing LLM-based methods in fault diagnosis primarily focus on improving classification accuracy through modality alignment or domain-specific fine-tuning. The question of how to effectively synthesize multi-source diagnostic evidence, including feature attributions, temporal trends, and confidence indicators, into operator-level interpretable reasoning remains largely unexplored. Meanwhile, ensuring the factual reliability of LLM-generated content in safety-critical manufacturing settings presents an additional engineering challenge that requires careful architectural consideration [27].

To address these challenges, this paper proposes a two-layer architecture that decouples diagnostic computation from interpretable reasoning through a structured evidence interface. The first layer is an evidence-augmented diagnostic engine that processes triaxial vibration signals through multi-domain feature extraction, PCA-Loading-based feature selection, and GRU-based temporal classification with IPSO-SA hyperparameter optimization. The second layer is an LLM-based reasoning layer that receives pre-computed classification results, feature attributions, temporal trends, and cross-model review flags through the evidence interface, and synthesizes them into coherent multi-evidence interpretations with context-adaptive action suggestions.

The main contributions of this work can be summarized in three aspects. First, the proposed two-layer architecture provides a structured approach to coupling deep learning-based wear classification with LLM-assisted interpretive reasoning, where factual content is confined to pre-computed fields to help maintain reliability in safety-critical manufacturing settings. Second, a Pareto-optimal diagnostic configuration is identified on the PHM2010 dataset. It reaches 98.38% macro F1-score with 10 of 84 original features. Third, blind evaluation by experienced CNC engineers against a basic and an enhanced rule template separates the value of evidence organization from that of LLM generation, with consistent factual accuracy confirmed across three LLMs and cross-dataset testing on NASA Ames milling records with different workpiece materials, sensor configurations, and cutting conditions providing preliminary evidence of generalizability.

The remainder of this paper is organized as follows. Section 2 reviews related work on data-driven tool wear monitoring, explainable AI in PHM, and LLM applications in industrial fault diagnosis. Section 3 describes the proposed methodology. Section 4 presents experimental results and discussion. Section 5 concludes the paper and outlines directions for future work.

# 2. Related work

## 2.1 Data-driven tool wear monitoring

Tool wear monitoring methods are generally categorized into direct and indirect approaches [3,28]. Direct methods use optical sensors or machine vision to measure geometric wear on the tool surface. While accurate, they require interrupting the machining process and are susceptible to interference from cutting fluids and chips, limiting their applicability in continuous production [29,30]. Indirect methods infer tool condition from process signals such as vibration, cutting force, acoustic emission, and spindle current acquired during machining, making them more suitable for online monitoring [31,32]. Recent work has demonstrated tool wear estimation in Inconel 718 drilling using spindle motor current with Bayesian state-space modeling, avoiding additional sensor installation [33].

With the advancement of sensor technology and deep learning, indirect methods based on multi-sensor signal analysis have become a widely adopted approach. Multi-sensor fusion frameworks have been developed for tool wear prediction, demonstrating the benefit of combining heterogeneous signal sources [34]. Raw multisensory signals have been fused through reshaped time-series CNNs to improve prediction accuracy [35]. Adaptive feature-temporal weighted methods combined with LSTM have also been proposed for capturing temporal wear evolution [36]. These studies demonstrate the effectiveness of deep learning in modeling complex nonlinear relationships between sensor signals and wear states.

Feature engineering remains important even when deep learning models are employed. Multi-domain feature extraction combining time-domain statistics, frequency-domain spectral descriptors, and time-frequency representations from wavelet packet decomposition has been shown to be effective for representing the diverse characteristics of progressive wear [9,37]. Deep forest approaches with parameter adaptive fine-tuning have been proposed for tool wear recognition under variable sensor combinations, highlighting the importance of robust feature representations [38].

Several recurrent architectures have been applied to model the temporal evolution of tool wear. A hybrid NCA-SMA-GRU model was developed for continuous wear prediction, using a metaheuristic algorithm to optimize GRU hyperparameters [4]. A deep heterogeneous GRU model was introduced for tool health monitoring through systematic feature engineering and hyperparameter search [5]. CNN-BiLSTM models with improved particle swarm optimization have also been proposed for tool wear prediction [10], and parallel CNN combined with BiLSTM has been applied for tool condition monitoring, though with manually configured hyperparameters [6]. Chaotic PSO has been used to optimize CNN-BiLSTM attention models [11], and a genetic PSO algorithm was developed to integrate feature fusion and hyperparameter optimization into a unified process, though without explicit escape mechanisms for local optima [8]. More recently, Transformer-based architectures have been explored for tool wear tasks, including temporal-frequency enhanced Transformers [7] and spatial-temporal adaptive graph convolutional networks [39]. Other recent contributions include stacked sparse autoencoders with ensemble learning [40], digital twin-driven online monitoring [41], and optimized sampling strategies for imbalanced data in tool wear monitoring [42]. Hierarchical temporal-modal fusion with uncertainty quantification has been proposed for tool wear recognition under multimodal monitoring data [43]. Simultaneous prediction of tool wear condition and surface roughness under limited samples has also been investigated [44], and generative image augmentation has been applied to address few-shot tool wear detection [45].

Despite the substantial progress in diagnostic accuracy, the outputs of these models remain numerical predictions or abstract representations. The gap between model output and practical decision support has received comparatively little attention in the tool wear monitoring literature. Comprehensive reviews have noted that interpretability remains a notable barrier to industrial adoption of AI-based tool monitoring systems [1,2]. A recent review further confirmed that integrating physics-based models with machine learning remains an active research direction for improving prediction reliability [46]. This observation motivates the development of reasoning mechanisms that can convert model outputs into actionable diagnostic guidance for practical replacement decisions.

## 2.2 Explainable AI in fault diagnosis and condition monitoring

Explainable AI methods have been increasingly applied to fault diagnosis to improve the transparency of data-driven models. The most widely used approaches are post-hoc attribution methods, particularly SHAP and LIME, which quantify the contribution of each input feature to a specific model prediction [12,13].

SHAP has been integrated with adversarial autoencoders for fault diagnosis of industrial processes, demonstrating that Shapley values can identify the process variables most responsible for detected anomalies [12]. Systematic comparisons of LIME, SHAP, Partial Dependence Plots, and Individual Conditional Expectation for predictive maintenance of rotating machines have provided comprehensive evaluations of model-agnostic explanation methods [13]. In the specific context of tool wear monitoring, lightweight and interpretable frameworks employing SHAP analysis have been proposed to quantify feature contributions while maintaining low computational overhead for edge deployment [14]. A systematic review of XAI methods for industrial fault diagnosis concluded that SHAP and feature importance methods are the most widely adopted, but noted significant gaps in outcome-level and end-user explainability [15].

These studies share a common characteristic from the perspective of operational deployment. The outputs of SHAP, LIME, and similar methods are numerical attribution scores or visual plots designed for model developers and data scientists. Converting a SHAP waterfall plot into a concrete tool replacement recommendation requires domain expertise that may not be readily available in typical machining environments. Furthermore, several of these methods involve stochastic sampling procedures that can produce slightly different attribution results across repeated evaluations of the same input, which may raise reproducibility considerations in safety-critical applications where deterministic outputs are preferred. The challenge is therefore not to replace these attribution methods but to develop a downstream reasoning mechanism that can convert their outputs into interpretable guidance for practical decision support.

## 2.3 Large language models in fault diagnosis and PHM

The emergence of large language models has opened new possibilities for translating numerical diagnostic outputs into human-interpretable guidance. Recent surveys have mapped the growing landscape of LLM applications in manufacturing and PHM [21–23].

A prominent research direction involves adapting LLMs to perform fault diagnosis directly. Engineering time-series data have been integrated into LLM fine-tuning through multimodal data alignment, enabling fault classification across different equipment types [16]. In manufacturing quality assessment, deep learning classifiers have been coupled with LLM-based analytical reasoning for resistance spot welding evaluation [47], and vision-language models have been applied to interactive quality assessment in laser powder bed fusion [48]. Domain-specific lightweight LLMs have been developed for HVAC fault detection and diagnosis [18], and adaptive industrial LLMs have been proposed for mechanical fault diagnosis under variable operating conditions [17]. Semantic-guided frameworks that bridge time-series and textual data with language models have been proposed for label-free multimodal fault diagnosis [19]. Multimodal data-enabled large models have also been developed for machine fault diagnosis towards intelligent operation and maintenance [49]. These approaches leverage the representational power of LLMs to improve diagnostic accuracy and cross-domain generalization.

A second direction focuses on integrating structured knowledge or domain constraints with LLMs to provide more reliable and interpretable diagnostic outputs. Knowledge graph-driven approaches combining improved joint extraction with large language models have been applied to aviation equipment fault diagnosis [24]. Physics-constrained multimodal LLMs have been proposed for fault diagnosis under imbalanced and under-sampled data conditions, demonstrating the value of incorporating domain constraints into LLM-based systems [25]. Explainable multimodal frameworks using large vision-language models have been developed for aviation hydraulic pump fault diagnosis, integrating visual and textual modalities [20]. Trustworthy multi-agent LLM frameworks incorporating physics-grounded tool orchestration and auditable reasoning provenance have been proposed for label-free bearing fault diagnosis [26].

Despite these advances, these methods mainly use LLMs to improve classification accuracy. The question of how to effectively leverage LLMs for multi-evidence reasoning, where multiple diagnostic outputs such as feature attributions, temporal trends, and confidence indicators are integrated into coherent, practically interpretable guidance, remains relatively underexplored. Current approaches either use LLMs as classifiers or employ them to process raw data directly, without a structured mechanism for grounding LLM reasoning in pre-computed diagnostic evidence. Meanwhile, LLMs are known to occasionally generate plausible but unsupported content in technical contexts [27], which presents an additional engineering challenge when deploying such models in safety-critical manufacturing environments.

This study addresses these gaps by proposing a two-layer architecture that separates diagnostic computation from interpretable reasoning. Unlike methods that fine-tune LLMs to perform diagnosis directly, the proposed approach assigns the LLM a reasoning role, where all factual content is pre-computed by a deterministic diagnostic engine and provided to the LLM through a structured evidence interface. The LLM then performs multi-evidence reasoning to produce practically interpretable outputs with context-adaptive action suggestions. This design aims to combine the reliability of deterministic diagnostic computation with the reasoning and communication capabilities of LLMs for practical decision support in tool wear management.

# 3. Methodology

## 3.1 Framework overview

The proposed framework decouples tool wear monitoring into two functionally independent layers connected by a structured evidence interface (Fig. 1).

The first layer is the Evidence-Augmented Diagnostic Engine. Raw triaxial vibration signals acquired during CNC milling are processed through a deterministic pipeline of multi-domain feature extraction (Section 3.2.1), feature selection (Section 3.2.2), temporal sequence construction (Section 3.2.3), and classification by a deep learning model (Section 3.2.4) whose hyperparameters are optimized via meta-heuristic search (Section 3.2.5). Unlike a standard classifier, this layer outputs not a bare label but a structured evidence record containing the predicted wear state, model confidence, per-sample feature attribution, temporal trend indicators, and a cross-model review flag.

The second layer is the LLM-based Multi-Evidence Reasoning Layer. An LLM-based reasoning layer receives the structured evidence and performs multi-evidence reasoning to produce practically interpretable outputs. The layer integrates classification results, feature attributions, temporal trends, and review flags into coherent interpretations, provides context-adaptive action suggestions, and adjusts communication urgency according to wear severity.

The interface between the two layers is a JSON evidence record containing the predicted wear state, model confidence, an independent Random Forest prediction, a review flag, an action code, and attribution and trend data for the five most influential features. The full field specification is detailed in Section 3.3.1 and Appendix B. This interface establishes an information boundary in which the LLM does not access raw vibration signals, model weights, or intermediate computational states. All diagnostic computation is completed within Layer 1, and Layer 2 focuses on reasoning over the pre-computed evidence to generate practically interpretable outputs.

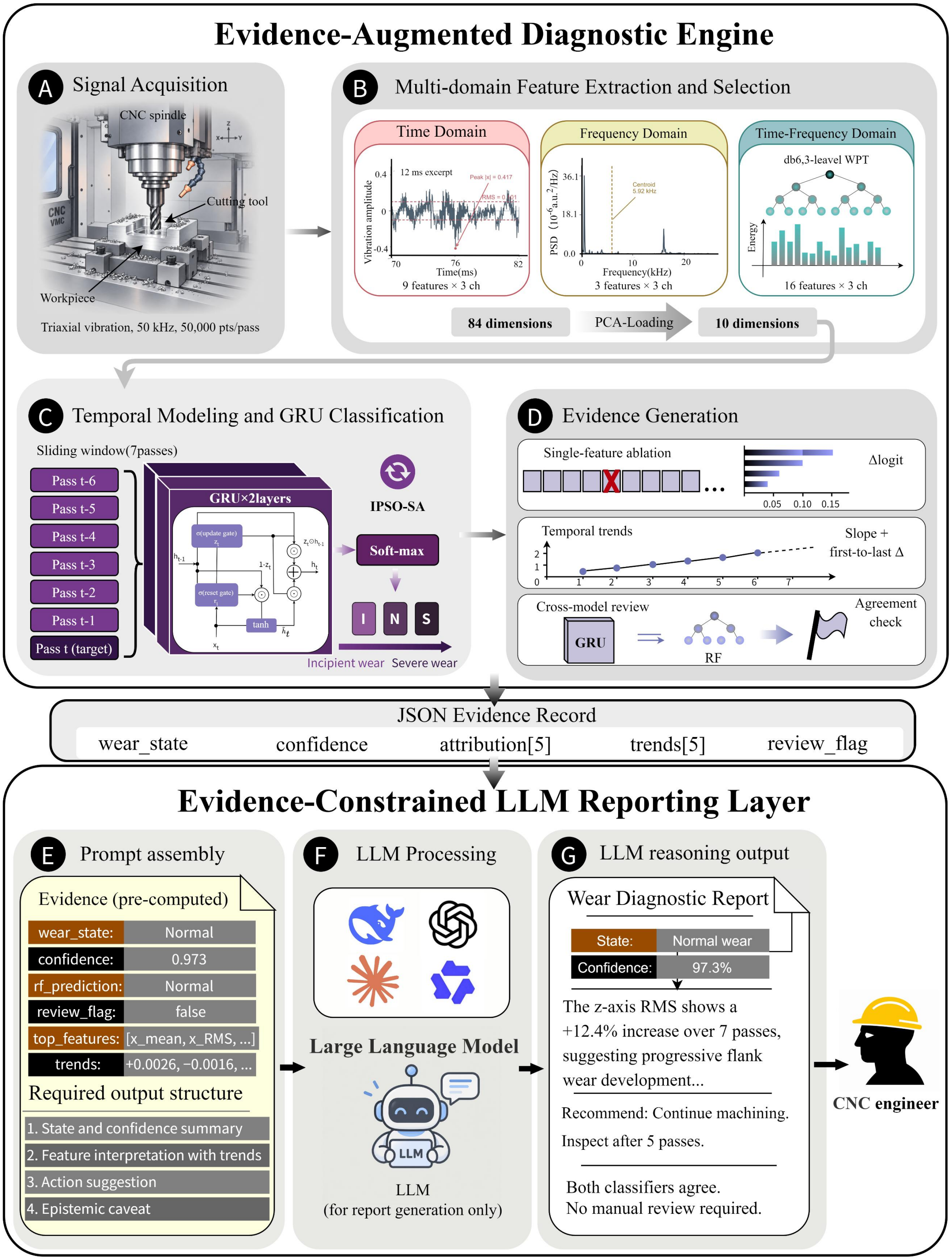

**Fig. 1.** Two-layer architecture for tool wear monitoring with multi-evidence reasoning.

## 3.2 Evidence-augmented diagnostic engine

### 3.2.1 Multi-domain feature extraction

Triaxial vibration signals (x, y, z) are captured by three accelerometers during milling. For each cutting pass, a fixed-length segment of 50,000 sampling points is extracted from the stable machining interval to avoid transient effects during tool entry and exit. Features are extracted from three complementary domains for each vibration channel independently.

Nine time-domain features capture the amplitude distribution and waveform characteristics of the vibration signal: mean, standard deviation, root mean square (RMS), peak absolute value, peak-to-peak value, kurtosis (Pearson's definition, computed as the ratio of the fourth central moment to the square of the second central moment), skewness (the ratio of the third central moment to the 1.5th power of the second central moment), shape factor (RMS divided by mean absolute value), and impulse factor (peak divided by mean absolute value). Among these, kurtosis is particularly sensitive to impulsive components associated with tool chipping, while RMS reflects overall vibration energy.

Three frequency-domain features characterize the global spectral shape. The power spectral density is estimated via the squared magnitude of the one-sided discrete Fourier transform. From this spectrum, three descriptors are computed: spectral centroid (power-weighted mean frequency), spectral RMS frequency (square root of the power-weighted mean squared frequency), and spectral variance (power-weighted variance around the centroid). These features describe the overall energy distribution across the frequency axis. As flank wear progresses, the increasing tool-workpiece contact area and rising cutting forces tend to shift spectral energy toward lower frequencies, making these descriptors sensitive to wear-induced changes in cutting dynamics.

Sixteen time-frequency domain features are extracted using three-level Wavelet Packet Transform (WPT) with the Daubechies-6 (db6) wavelet basis. This decomposes each channel into $2^3 = 8$ equal-bandwidth sub-bands. Given the 50 kHz sampling frequency (Nyquist frequency: 25 kHz), the sub-bands span [0, 3125] Hz through [21875, 25000] Hz in 3125 Hz increments. Five candidate wavelet bases (db4, db6, db8, sym6, coif3) were evaluated under identical GRU classification settings. The five bases gave similar macro F1-scores in a preliminary test, which shows low sensitivity to basis selection. The db6 basis was adopted.

For each sub-band j ($j$=1,…,8), the energy is computed from the wavelet packet coefficients $c_{j,n}$, and the energy fraction is obtained by normalization:

$$p_j=\frac{E_j}{\sum_{i=1}^{8} E_i}, \quad E_j=\sum n\,|c_{j,n}|^2 \tag{1}$$

The wavelet energy entropy for each sub-band is then defined as:

$$H_j=-p_j \ln(p_j) \quad (p_j>0) \tag{2}$$

and the entropy share is computed as:

$$S_j=\frac{H_j}{\sum_{i=1}^{8} H_i} \tag{3}$$

The 8 entropy values and 8 entropy shares yield 16 time-frequency features per channel. Wavelet packet entropy captures the complexity of energy distribution across frequency sub-bands, which tends to increase as wear-induced broadband vibration components develop during tool degradation. Concatenating all three domains across three vibration axes produces 3×(9+3+16)=84 features per cutting pass.

3.2.2 Feature selection

The 84-dimensional feature space contains redundant and weakly discriminative features that may degrade classification performance, particularly with limited training samples. Rather than applying a fixed dimensionality reduction, three feature selection methods are evaluated systematically to identify the most effective subset.

Random Forest importance ranks features by their Gini impurity-based importance scores obtained from a Random Forest classifier trained on the standardized training features. Mutual Information (MI) estimates the mutual information between each individual feature and the wear state label on the training set, ranking features by decreasing MI score. This method captures nonlinear statistical dependencies without assuming a specific model structure.

PCA-Loading leverages the loading matrix of Principal Component Analysis to quantify each original feature's contribution to the dominant variance directions. PCA is fitted on the standardized training feature matrix $\mathbf{Z} \in \mathrm{R}^{n\times 84}$. The eigenvalue decomposition of the correlation matrix $\mathbf{R}=\frac{1}{n-1}\mathbf{Z}^{\top}\mathbf{Z}$ yields eigenvalues $\lambda_1 \geq \lambda_2 \geq \cdots \geq \lambda 84$ and corresponding eigenvectors (loadings) $\mathbf{w}i \in \mathrm{R}^{84}$. Each principal component PC$i$=$\mathbf{Z}\mathbf{w}i$ is a linear combination of all 84 original features, with the explained variance ratio:

$$\eta_i=\frac{\lambda_i}{\sum_{m=1}^{84} \lambda_m} \tag{4}$$

The first $q$ components are retained such that the cumulative explained variance $\sum_{i=1}^{q} \eta_i$ exceeds 95%. Rather than projecting the data onto these $q$ principal component axes, which would produce abstract linear combinations without physical meaning, the loading matrix is used to quantify each original feature's contribution to the retained variance. A composite importance score for each original feature $j$ is computed as:

$$I_j = \sum i=1^q \eta_i \cdot \ wij^2 \tag{5}$$

where $w_{ij}$ is the loading of feature $j$ on component $i$. Features are ranked by decreasing $I_j$. This score reflects how strongly each physically interpretable original feature participates in the dominant variance directions of the data, without conflating principal components with individual features. Unlike PCA-based dimensionality reduction, PCA-Loading selects a subset of original features, maintaining direct physical interpretability.

For a given selection method, the top-k features are retained, where k is treated as a tunable hyperparameter evaluated jointly with the sequence length N (Section 3.2.3).

3.2.3 Temporal sequence construction

Tool wear is a cumulative degradation process: the current wear state depends not only on the present cutting pass but also on the preceding trajectory. To provide the classifier with temporal context, a sliding-window approach constructs input sequences of N consecutive passes. For a target pass at index t, the input sequence comprises passes $[t\text{-}N+1, t\text{-}N+2, \ldots, t]$, and the label corresponds to the wear state of pass $t$. Sequences are constrained to never cross tool boundaries: if pass $t$ is within the first $N$-1 passes of a given tool, no sequence is formed with pass $t$ as the target. The resulting input tensor has shape $(B,N,k)$, where $B$ is the number of valid sequences and $k$ is the number of selected features. Prior to model input, features are standardized using z-score normalization fitted exclusively on the training set. The sequence length N is evaluated at N ∈ {3,5,7,10} jointly with the feature count k. To ensure that performance differences across N values are not confounded by different target-pass sets, all four sequence lengths share a common target set restricted to passes 10 through 315 per tool, yielding 3 × 306 = 918 valid samples regardless of N.

3.2.4 Classification architectures

Six sequence classification architectures are evaluated to identify the most suitable model for this small-sample PHM task.

GRU (Gated Recurrent Unit) employs two stacked layers that process the temporal sequence through an update gate $z_t$ and a reset gate $r_t$. The core state update equations are:

$$z_t=\sigma(Wxzx_t+Whzh_{t-1}) \tag{6}$$

$$r_t=\sigma(Wxrx_t+Whrh_{t-1}) \tag{7}$$

$$\tilde{h}t=\tanh(Wx_t+r_t \odot Uht-1) \tag{8}$$

$$h_t=z_t \odot ht-1+(1-z_t) \odot \tilde{h}t \tag{9}$$

where $x_t$ denotes the input feature at the current time step, $h_{t-1}$ is the hidden state from the previous step, $\sigma$ is the sigmoid activation function, and $\odot$ indicates element-wise multiplication. The hidden state of the final time step is passed to a linear classification head. GRU's simplified gating mechanism yields fewer parameters than LSTM at the same hidden size, which is potentially advantageous when training data are limited.

LSTM (Long Short-Term Memory) uses two stacked layers with the standard forget-input-output gate structure; Bi-LSTM extends this with bidirectional processing, concatenating forward and backward hidden states before the classification head. 1D-CNN applies two convolutional layers (kernel size 3) with ReLU and dropout, followed by adaptive average pooling and a linear classifier.

Transformer uses a standard encoder with learned positional embeddings, two encoder layers, four attention heads, and GELU activation. PatchTST is a classification-adapted channel-independent Transformer variant with length-3 stride-2 patches and a shared two-layer encoder; RevIN normalization is omitted to preserve absolute feature levels.

All six architectures share the same training protocol: Adam optimizer, cross-entropy loss function as defined in Eq. (10), and identical train/test splits. The loss function is:

$$L=-\sum_{c=1}^{C} y_c \log P(y_c) \tag{10}$$

where $C$ denotes the number of wear state categories, $y_c$ represents the classification label, and $P(y_c)$ is the predicted probability. Hyperparameters (hidden size, dropout rate, learning rate, number of epochs, batch size) are optimized independently for each architecture using the procedure described in Section 3.2.5.

3.2.5 Hyperparameter optimization

The classification performance of deep learning models is sensitive to hyperparameter configuration. Six optimization methods are compared under equal computational budgets to identify the most effective strategy. Five hyperparameters are jointly optimized: hidden layer size $\in[16,256]$, dropout rate $\in[0.1,0.5]$, number of training epochs $\in[10,100]$, batch size $\in[16,128]$, and learning rate $\in[10^{-4},10^{-2}]$ (log-uniform). All coordinates are internally normalized to [0,1] for algorithm-agnostic handling, with learning rate decoded logarithmically.

The six methods are: Random Search (uniform sampling), Bayesian Optimization via Tree-structured Parzen Estimator (BO-TPE) with 10 random startup trials, Simulated Annealing (SA) with a single-solution trajectory and cooling schedule $T$=0.1×0.001$^{t/t_{\max}}$, standard Particle Swarm Optimization (PSO) with constriction coefficients, Genetic Algorithm (GA) with tournament selection and arithmetic crossover, and the proposed Improved PSO with Simulated Annealing (IPSO-SA).

The IPSO-SA algorithm extends standard PSO with two mechanisms. The first is dynamic inertia weight adjustment, where ω decreases linearly from 0.9 to 0.4 as specified in the parameter configuration below. The second is an SA-based acceptance mechanism whereby a particle whose updated position yields inferior fitness is not automatically rejected but accepted with the Metropolis probability defined in Eq. (13), allowing occasional traversal of inferior regions to reduce premature convergence.

Both PSO and IPSO-SA use 10 particles with zero initial velocity and velocity clamping to [−0.2, 0.2] in normalized coordinates, with c1 = c2 = 1.49618. Standard PSO uses a fixed inertia weight ω = 0.7298 following the Clerc-Kennedy constriction factor. IPSO-SA replaces this with a linearly decreasing ω from 0.9 to 0.4, promoting global exploration in early iterations and local refinement in later stages. The velocity and position updates for particle $i$ in dimension $d$ follow:

$$vid^{t+1}=\omega vid^{t}+c_1r_1(pid^{t}-xid^{t})+c_2r_2(g_d^t-x_{id}^t) \tag{11}$$

$$xid^{t+1}=xid^{t}+v_{id}^{t+1} \tag{12}$$

where $p_{id}$ is the personal best position, $g_d$ is the global best, $r_1,r_2\sim U(0,1)$, and velocities are clamped to [-0.2,0.2] in normalized coordinates. When a particle's updated position is inferior to its personal best, it replaces the personal best with probability:

$$p=exp\left(\frac{\Delta f}{T}\right) \tag{13}$$

where $\Delta f$ is the fitness difference and $T$ is the current temperature following the cooling schedule. The best configuration found during the search is retained separately and returned as the final result. Each optimization method receives an identical budget of fitness evaluations per seed, with macro-averaged F1-score on the validation set as the fitness function.

## 3.3 LLM-based multi-evidence reasoning layer

The diagnostic engine outputs a predicted wear state, a softmax probability vector, and a set of per-feature attribution scores. These outputs are sufficient for algorithmic evaluation but are not readily interpretable for practical tool replacement decisions. To address this, an LLM-based reasoning layer is introduced to perform multi-evidence reasoning over the diagnostic outputs, integrating multiple evidence sources into coherent, practically interpretable outputs with context-adaptive action suggestions. An information boundary is established such that all factual content is computed deterministically within the diagnostic engine before the LLM is invoked. The LLM does not access raw vibration signals, model weights, or intermediate states, and instead focuses on reasoning over the pre-computed evidence to support operator decision-making.

### 3.3.1 Evidence record generation

For each classified sample, the diagnostic engine generates a structured evidence record through three analyses.

Feature attribution via single-feature ablation. Single-feature ablation is chosen because it is simple and deterministic. It needs only k extra forward passes per sample, and every evidence field is exactly reproducible. Each of the k retained features is independently replaced by zero in the standardized input (equivalent to substituting the training-set mean), and the model is re-evaluated. The ablation delta for feature j is defined as:

$$\Delta_j = l_{\hat{y}}(\mathbf{x}) - l_{\hat{y}}(\mathbf{x}^{\setminus j}) \tag{14}$$

where $l_{\hat{y}}(\cdot)$ denotes the pre-softmax logit for the predicted class $\hat{y}$, $x$ is the original standardized input, and $x^{\setminus j}$ is the input with feature j replaced by zero. A positive $\Delta_j$ indicates that the feature supports the current classification; a negative value indicates it works against it. The five features with the largest $|\Delta j|$ are selected as top contributors. For sequential models, each feature is zeroed across all N time steps simultaneously; for single-pass classifiers, the ablation applies to the current-pass vector directly.

Temporal trend computation. For each top feature, two indicators are computed from the raw feature values within the N-pass window: the first-to-last difference and the slope of a least-squares linear fit. For single-pass classifiers without temporal windows, trend computation is omitted.

Cross-model review flag. A binary flag is raised when the maximum softmax probability falls below 0.85 or the primary classifier disagrees with an independent secondary classifier. The two classifiers use intentionally different input representations to maximize detection diversity: in the PHM2010 configuration, the GRU operates on a 7-step sequence of 10 selected features, while the RF operates on the current-pass 84-dimensional vector. The flag does not alter the classification but signals that human verification is recommended. The 0.85 threshold is exploratory and has not been calibrated for optimal sensitivity-specificity trade-off. An action code is then assigned deterministically from the predicted state and the review flag: REVIEW_NOW whenever the flag is raised, REPLACE for a severe prediction without a flag, and CONTINUE otherwise. The flag can therefore only raise the action level, and the LLM reports the code but cannot change it.

The three analyses are assembled into a JSON evidence record containing the predicted wear state, softmax probability, ablation deltas and ranks for the top five features, temporal trend indicators, the independent classifier prediction, the review flag, and the action code. The complete field specification is provided in Appendix B.

3.3.2 Prompt design

The LLM receives the evidence record and a prompt template specifying the required output structure: a summary of the wear state and confidence, interpretation of each top feature including trend behavior and classification influence, a recommended action consistent with the pre-computed action code. All numerical values and their derived categorical interpretations (trend direction, feature effect) are pre-computed and embedded in the prompt as factual constraints that the LLM is expected to preserve. The LLM contributes multi-evidence reasoning, context-specific interpretation, and targeted action suggestions. Two rule-based templates are constructed for comparison. The basic template (Condition D) inserts the same fields into fixed sentence patterns via string formatting. The enhanced template (Condition E) applies one rule per evidence dimension: it groups the top features by the sign of their ablation deltas, describes trends as sustained or reversing, states any classifier disagreement, and selects urgency wording from the action code, without rules for specific combinations of evidence. The full prompt template is provided in Appendix A.

3.3.3 Output evaluation methodology

The quality of LLM-generated outputs is evaluated along three dimensions.

Factual accuracy measures whether the output content is consistent with the evidence record. Two checks are applied: numerical fidelity verifies that every numerical value in the output matches the evidence record exactly, and factual consistency verifies that every qualitative assertion (trend directions, feature effects, wear state) is consistent with the underlying computed data. An output passes factual accuracy only if both checks are satisfied. An output is counted as fully faithful when it passes all three checks and reproduces the pre-computed action code. Both checks are performed by a script that parses each output, matches every reported number to the corresponding evidence field at the reported precision, and compares each categorical statement (wear state, trend direction, and feature effect) with the evidence record. To verify the script, two authors independently checked a random sample of 100 outputs, and their judgments agreed with the script in all cases.

Information completeness verifies that the output contains all elements required for a replacement decision: predicted wear state, confidence level, interpretation of the five top features, recommended action.

Practical decision support value assesses whether the output supports practical decision-making, measured through blind human evaluation. The LLM report and the two template reports for the same evidence record are presented in randomized order to evaluators with CNC machining experience, who rate each output on actionability, overall quality, clarity, and decision confidence using five-point Likert scales. Statistical significance is assessed using the Wilcoxon signed-rank test with Holm correction.

# 4. Experiments and Results

## 4.1 Datasets and experimental setup

Two publicly available datasets are used in this study. Their key characteristics are summarized in Table 1.

Table 1. Dataset overview.

| Item | PHM2010 | NASA Ames Milling |
|---|---|---|
| Machine | Röders Tech RFM760 | Matsuura MC-510V machining center |
| Workpiece | Stainless steel (HRC52) | Cast iron / stainless steel J45 |
| Tool | 3-flute ball-nose carbide | 70 mm face mill with six KC710 inserts |
| Spindle speed | 10,400 rpm | 826 rpm (cutting speed 200 m/min) |
| Feed rate | 1,555 mm/min | 0.254 / 0.508 mm/rev |
| Axial depth of cut | 0.2 mm | 0.75 / 1.5 mm |
| Sampling frequency | 50 kHz | 250 Hz |
| Signal channels | Vibration x, y, z | AC current, DC current, table vibration, spindle vibration, table AE, spindle AE |
| Labeled samples | 945 (3 tools × 315 passes) | 146 (16 cases, 21 records missing VB) |
| Wear measurement | Optical microscopy, per-pass | Optical microscopy, per-run |
| Role in this study | Primary validation | LLM reasoning layer cross-validation |

The PHM2010 dataset comprises three cutting experiments (C1, C4, C6), each consisting of 315 tool passes. Flank wear is recorded as the maximum value across three cutting edges, converted from micrometers to millimeters. For each pass, a fixed segment of 50,000 sampling points is extracted from the stable machining interval to avoid transient effects during tool entry and exit. Although PHM2010 also provides cutting force and acoustic emission signals, only the triaxial vibration channels are used, because accelerometers can be mounted on production machines at low cost without constraining the workpiece or fixture, whereas dynamometers are costly and limit the workpiece size. Under leave-one-tool-out evaluation (Table 6), adding force and acoustic emission channels did not improve cross-tool performance under the VB-based labels used in this study, although it substantially improved performance under pass-index stage labels. Neither dataset provides predefined wear-state labels. Wear states are defined from the measured flank wear rather than from pass-index intervals, because replacement decisions depend on the wear magnitude, whereas pass-index stages reflect cutting progression rather than measured wear. For PHM2010, three candidate VB threshold pairs were evaluated, and the 0.08/0.13 mm pair was selected for producing all three classes across every tool with the highest class-balance entropy. For NASA Ames, thresholds of 0.2/0.4 mm were set at the rounded 33rd and 67th percentiles of the empirical VB distribution. These thresholds are relative grades within the VB range of each dataset. They do not correspond to the ISO tool life criterion. The resulting class assignments are shown in Table 2.

Table 2. Class distribution.

| Dataset | Threshold | Incipient | Normal | Severe | Total |
|---|---|---|---|---|---|
| PHM2010 | VB < 0.08 / 0.08~0.13 / >= 0.13 mm | 100 | 558 | 287 | 945 |
| NASA Ames | VB < 0.2 / 0.2~0.4 / >= 0.4 mm | 50 | 44 | 52 | 146 |

The PHM2010 class distribution is imbalanced, with incipient wear at only 10.6% of total samples, making incipient-wear recall the most discriminative performance indicator throughout the experiments.

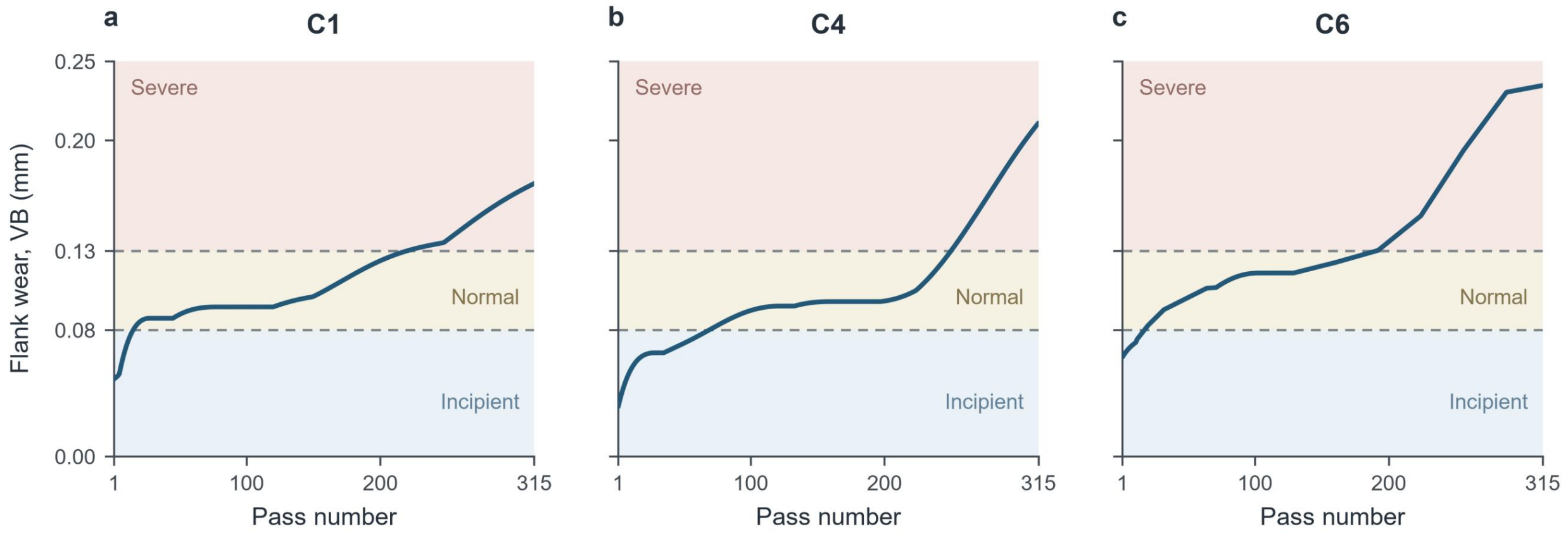

Fig. 2. Flank wear progression curves for experiments C1, C4, and C6.

Fig. 2 presents the flank wear progression curves for the three PHM2010 tools, illustrating the characteristic three-stage pattern of rapid initial wear, prolonged stable growth, and accelerating degradation. Differences in wear trajectories across cutting edges introduce nonlinear characteristics that challenge classification, particularly at state boundaries. To illustrate the signal-level differences across wear states, Fig. 3 presents triaxial vibration waveforms from three representative passes of tool C4, selected as the passes closest to the median VB value within each wear class. A fixed 20 ms window is displayed for all nine subplots using identical vertical axis limits to allow direct visual comparison. Vibration amplitude increases progressively from incipient to severe wear across all three channels, with the x-axis channel exhibiting the most pronounced amplitude growth. These signal-level changes motivate the use of amplitude-based time-domain features such as RMS and standard deviation as primary wear indicators.

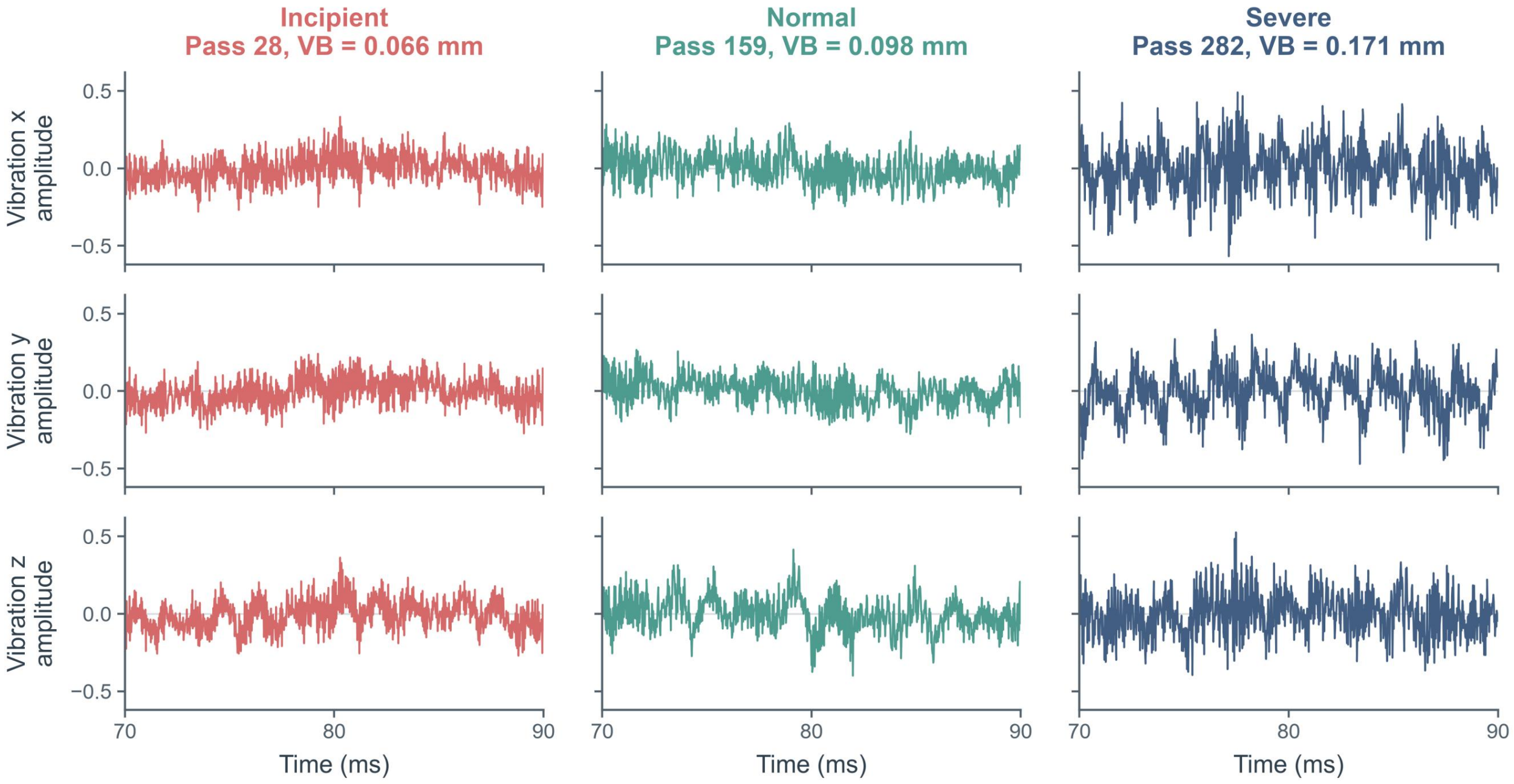

Fig. 3. Triaxial vibration waveforms across three wear states (C4, z-score normalized, 20 ms window). Each column corresponds to a representative pass selected as the closest to the median VB value within its wear class. All subplots share identical vertical axis limits.

Fig. 4 illustrates the multi-domain feature extraction process applied to the z-axis vibration signal of C4, Pass 159. The time-domain waveform (a) is annotated with two representative features, peak absolute value and RMS. The corresponding power spectral density (b) shows that the dominant spectral peaks lie below 5 kHz, while residual high-frequency components raise the spectral centroid to 5.92 kHz. The wavelet packet decomposition (c) distributes energy across eight sub-bands, with sub-band B1 containing 59.8% of the total energy and the remaining energy distributed among B2, B5, and B6. The 84 features used in subsequent analysis are constructed by applying this extraction process to all three vibration channels independently.

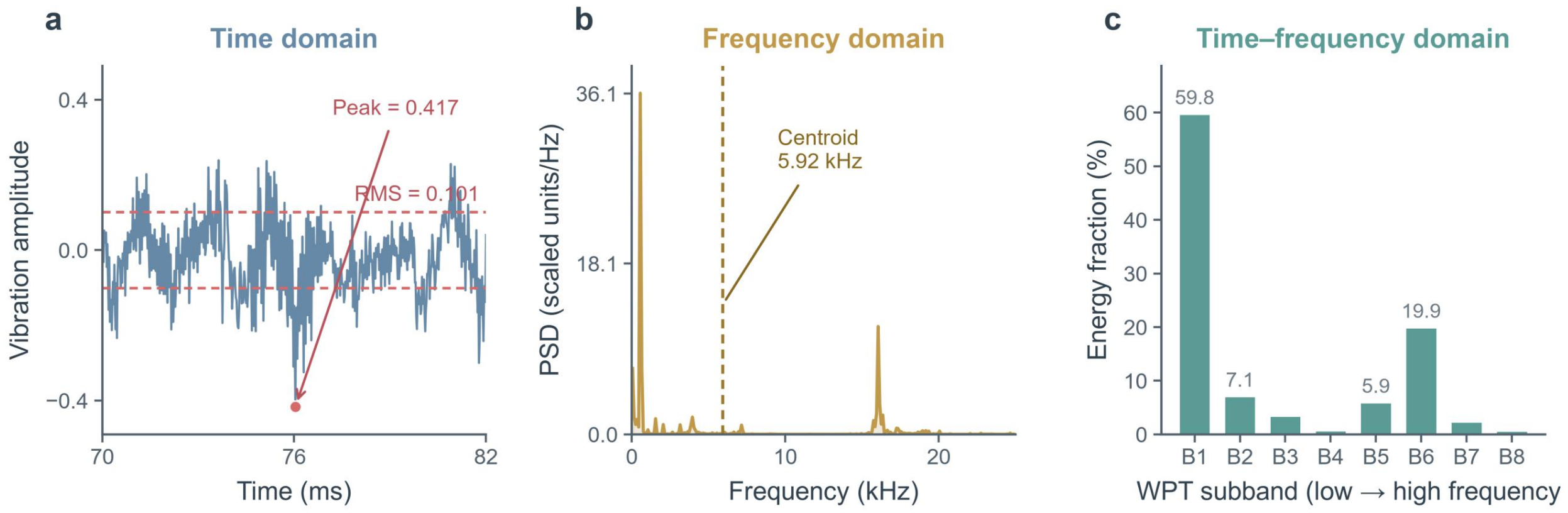

Fig. 4. Multi-domain feature extraction illustrated on the z-axis vibration signal of C4, Pass 159. (a) Time-domain waveform with peak and RMS annotations. (b) Power spectral density with spectral centroid. (c) Wavelet packet energy fraction across eight sub-bands from three-level db6 decomposition.

The NASA Ames Milling dataset serves as a secondary validation platform for the LLM reasoning layer rather than for the diagnostic engine. Its substantially different signal characteristics (250 Hz mixed-channel versus 50 kHz triaxial vibration) and small sample size (146 labeled records) provide a test of the reasoning layer's transferability across datasets, signal modalities, and classifier architectures.

All deep learning experiments are conducted using PyTorch 2.10.0 with CUDA on an NVIDIA RTX 5070 Laptop GPU. The evidence-constrained reasoning layer uses DeepSeek V4 Flash via API with temperature set to 0 to minimize output variability and maximum token length of 2048. The Random Forest classifier, used for both the NASA Ames diagnostic engine and the PHM2010 review flag, was trained on the current-pass feature vector (84 dimensions for PHM2010 and 168 for NASA Ames) with 100 trees, Gini impurity, unrestricted depth, bootstrap sampling, and square-root feature subsampling. Classification performance is evaluated using macro-averaged F1-score and per-class recall, reported as mean ± standard deviation over five random seeds (42–46), each defining a stratified 80/20 pass-level split. Within each training split, a stratified 20% subset was held out as the validation set for hyperparameter search, and the selected configuration was retrained on the full training split before a single evaluation on the test split. Because the split is made at the pass level and consecutive input windows overlap, a test window shares most of its passes with training windows from the same tool, and adjacent passes have nearly identical flank wear. The results in Tables 3–5 therefore measure recognition within the wear trajectories of the three tools, whereas cross-tool generalization is assessed separately by the leave-one-tool-out evaluation in Table 6.

## 4.2 Feature selection and sequence length optimization

The 84-dimensional multi-domain feature space described in Section 3.2.1 contains substantial redundancy. To determine the optimal feature subset and temporal context window, a joint sensitivity study was conducted across three feature selection methods (RF Importance, Mutual Information, PCA-Loading), nine feature counts (k = 3, 5, 8, 10, 15, 20, 30, 50, 84), and four sequence lengths (N = 3, 5, 7, 10), totaling 540 GRU training runs with fixed hyperparameters (hidden size 64, 2 layers, dropout 0.3, learning rate 0.001, 50 epochs, batch size 32).

Table 3 presents the PCA-Loading results across all evaluated feature counts at N = 7. Performance improves sharply from k = 3 to k = 8 (macro F1 from 87.04% to 97.95%), then stabilizes. The full 84-feature set achieves the highest macro F1 (98.81%), while k = 10 and k = 84 share the highest incipient recall (98.67%).

Table 3. PCA-Loading feature selection results at N = 7 with fixed-hyperparameter GRU.

| **k** | **Macro F1 (%)** | **Incipient Recall (%)** | **Features Retained** |
|---|---|---|---|
| 3 | 87.04 ± 4.37 | 86.67 | 3.6% |
| 5 | 96.34 ± 2.82 | 92.00 | 6.0% |
| 8 | 97.95 ± 0.29 | 94.67 | 9.5% |
| 10 | 98.47 ± 0.94 | 98.67 | 11.9% |
| 15 | 98.14 ± 0.87 | 97.33 | 17.9% |
| 20 | 98.11 ± 1.57 | 96.00 | 23.8% |
| 30 | 97.81 ± 2.00 | 93.33 | 35.7% |
| 50 | 97.89 ± 2.78 | 93.33 | 59.5% |
| 84 | 98.81 ± 1.17 | 98.67 | 100% |

Macro F1 is reported as mean ± sample standard deviation across five random seeds (42–46). Incipient recall is the mean across the same seeds. Features retained = k/84 × 100%. Each stratified test split contained approximately 15 incipient-wear samples, so a single additional misclassification shifts incipient recall by 6.67 percentage points.

The selection of k = 10 is based on a Pareto trade-off between feature compression and classification performance. From k = 5 to k = 10, macro F1 gains 2.13 percentage points and incipient recall gains 6.67 percentage points. Beyond k = 10, marginal returns diminish: macro

F1 increases only 0.34 percentage points from k = 10 to k = 84 while requiring 8.4 times as many features. Notably, k = 10 matches the incipient recall of the full 84-feature set (both 98.67%), indicating that the dominant discriminative information for the most challenging class is concentrated in a small number of features. Incipient recall at k = 15 and k = 20 is slightly lower than at k = 10. This difference equals about one test sample per split and is within noise. The k = 10 configuration achieves 99.66% of the peak macro F1 while retaining only 11.9% of the original feature space.

Fig. 5 shows the variance-weighted loading scores of the 10 selected features, ranked by their mean composite score across five random seeds. Eight of the 10 features are time-domain amplitude statistics, namely the mean and standard deviation of all three axes and the RMS of the x and y axes. The remaining two are the z-axis spectral RMS frequency from the frequency domain and the z-axis wavelet packet entropy in sub-band 7 from the time-frequency domain. The three mean vibration features rank highest. The z-axis wavelet packet entropy feature exhibits the largest cross-seed variance, suggesting that its contribution to the dominant variance directions is less stable than that of the time-domain features.

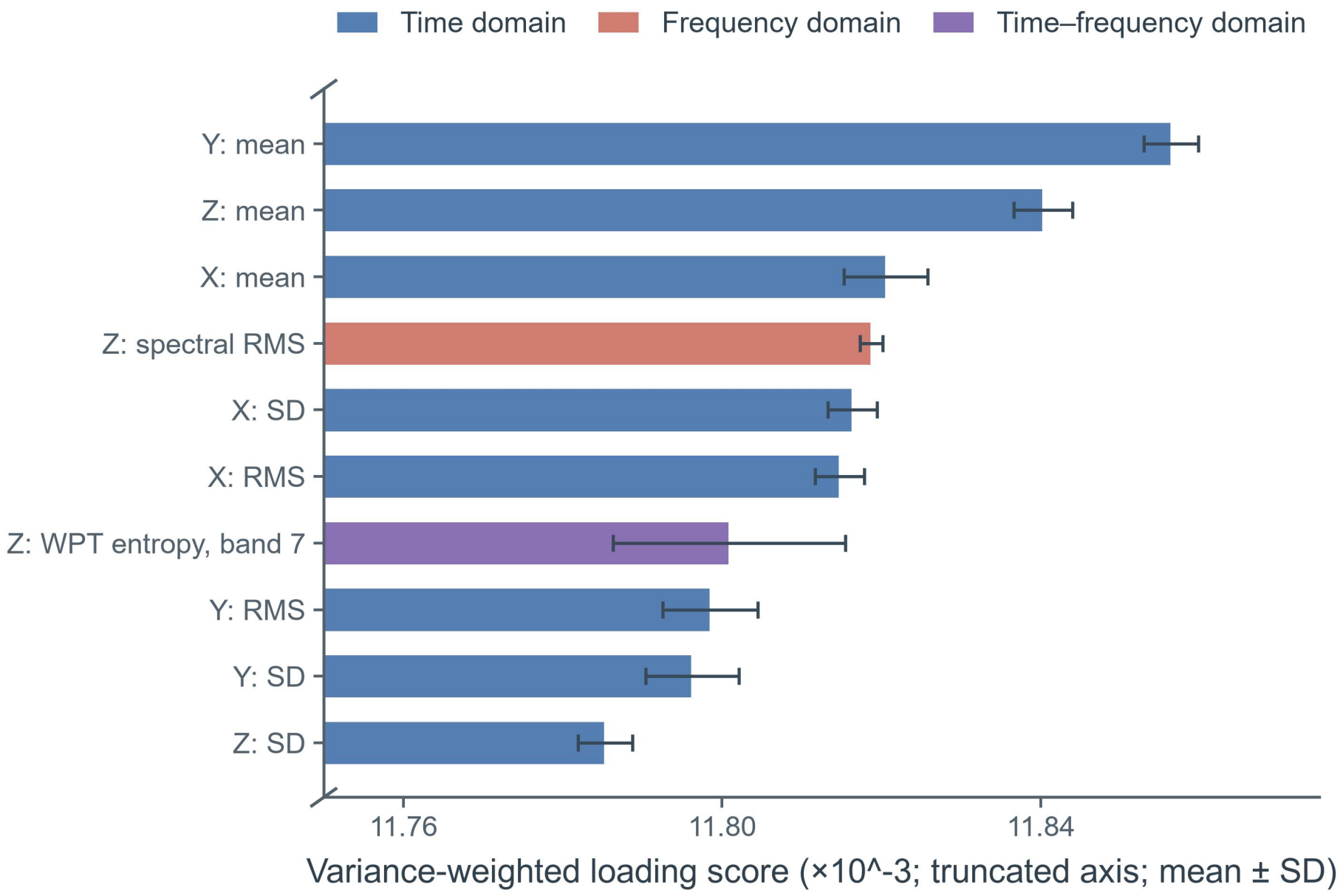

Fig. 5. Variance-weighted loading scores of the 10 features selected by PCA-Loading, ranked by mean composite score across five random seeds. Error bars indicate one standard deviation. Colors distinguish time-domain, frequency-domain, and time-frequency-domain features.

PCA-Loading is selected for its advantage at low feature counts. At k = 3, PCA-Loading achieves 86.67% incipient recall versus 45.33% (RF Importance) and 46.67% (Mutual

Information). PCA-Loading is unsupervised, so this gap is an empirical observation for this dataset. Although the gap narrows at larger k, PCA-Loading retains a substantial advantage at the selected operating point: at k = 10 and N = 7, PCA-Loading achieves 98.67% incipient recall versus 80.00% (RF Importance) and 81.33% (Mutual Information), which supports its use at the selected operating point.

Fig. 6 presents the incipient-wear recall heatmap across all k, N, and selection method combinations. The heatmap reveals that performance saturates around k = 10 for all three methods. At the selected operating point of k = 10, N = 7 and N = 10 achieve nearly identical macro F1 scores (98.47% versus 98.53%); the shorter window is adopted as the more parsimonious configuration.

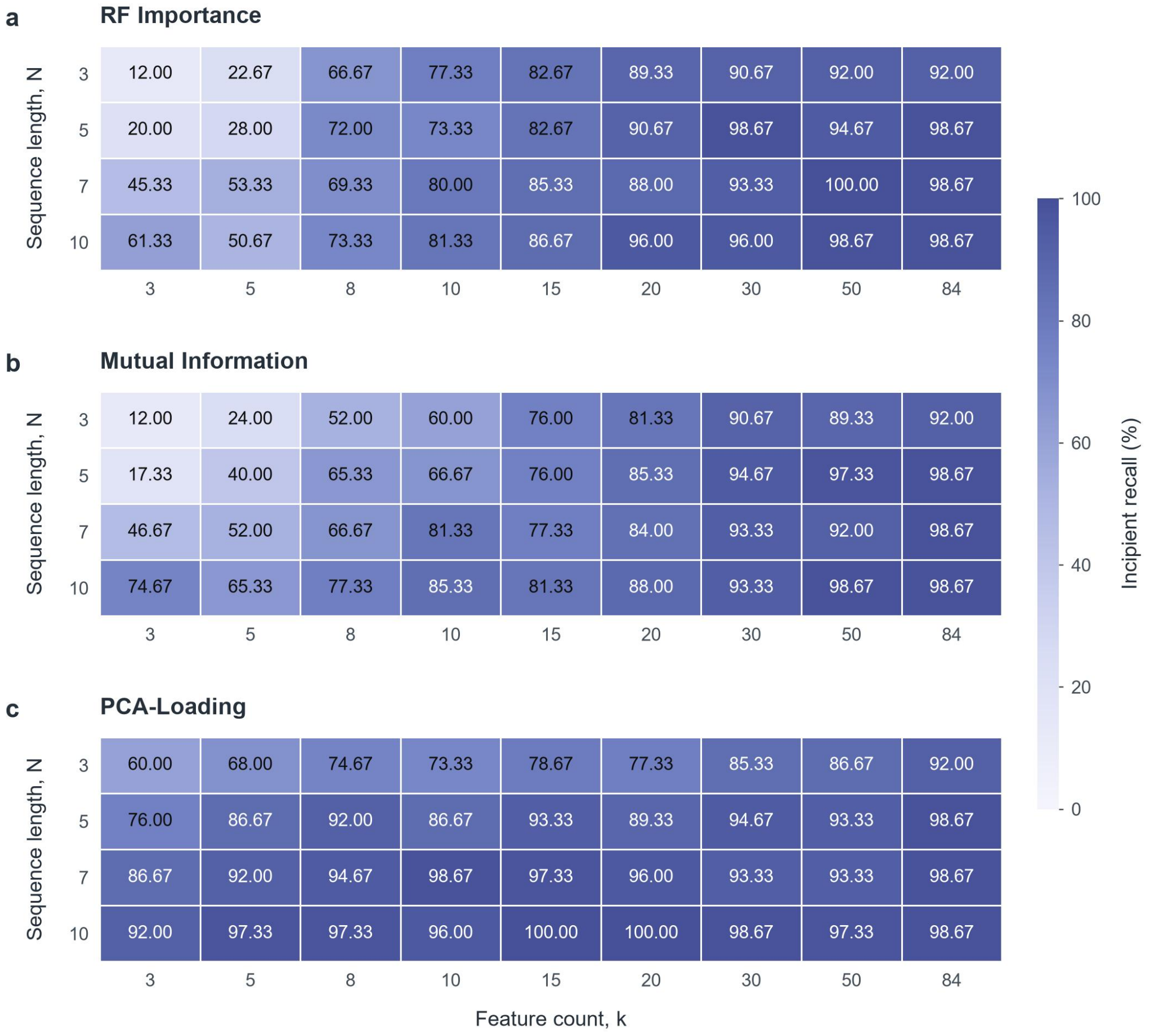

Fig. 6. Incipient-wear recall heatmap across k, N, and feature selection methods.

To verify that the selected 10-feature subset preserves class separability, Fig. 7 compares the PCA projections of the full 84-feature space (a) and the selected 10-feature space (b). In the full feature space, the first two principal components explain 57.6% of the total variance, and the three wear classes overlap substantially in the central region. In the selected feature space, the first two components explain 83.9% of the variance, and the class boundaries become somewhat more distinct. The two-dimensional silhouette score increases from 0.068 to 0.172. Both values are low, but the trend suggests that the selected features keep class separability.

Based on these results, the configuration PCA-Loading with k = 10 and N = 7 is adopted for all subsequent experiments.

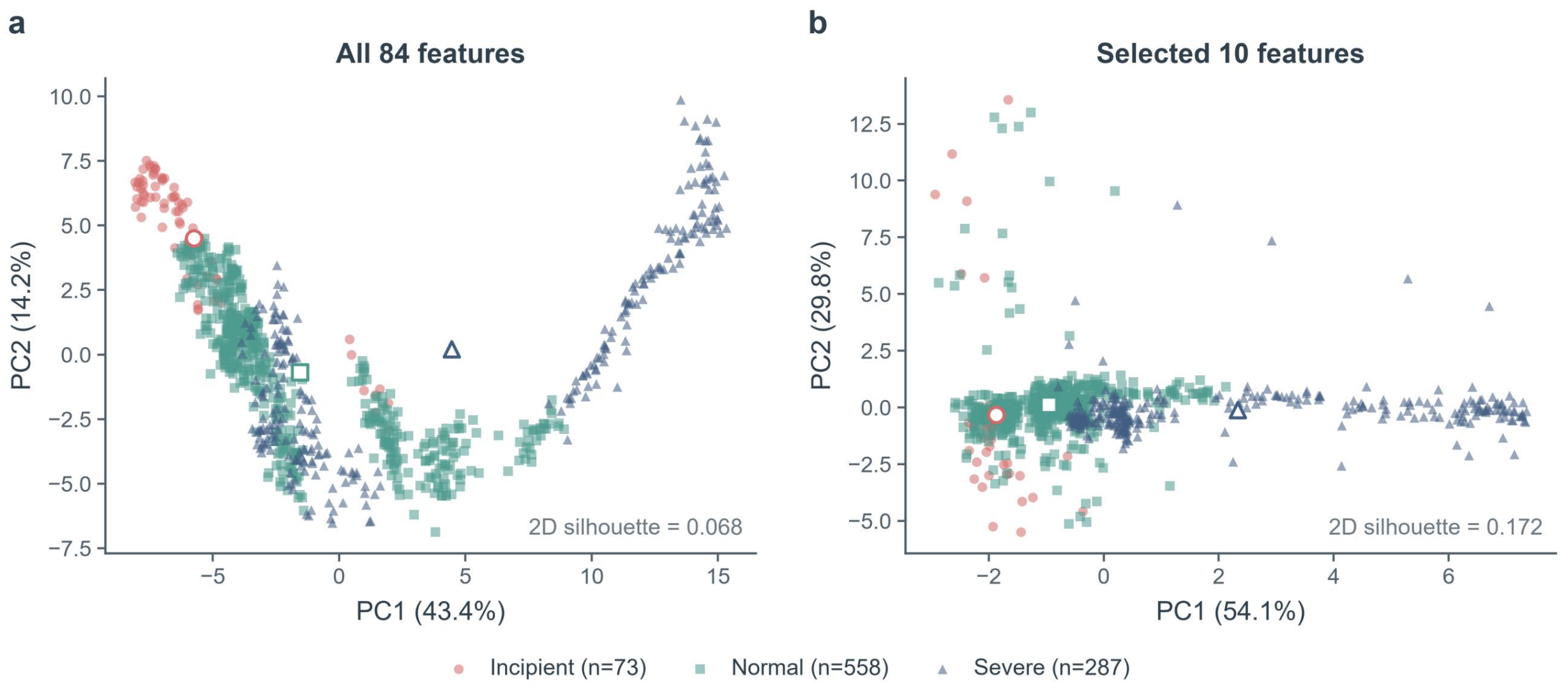

Fig. 7. PCA projection of the feature space using all 84 features (a) and the selected 10 features (b). Each point represents one target pass under the common-target protocol (passes 10 to 315, n = 918; 27 incipient passes from passes 1 to 9 are excluded to maintain a consistent target set across sequence lengths). Open markers indicate class centroids. Two-dimensional silhouette scores are annotated in each subplot.

## 4.3 Model architecture and optimization algorithm comparison

To identify the most suitable classification architecture, six models were evaluated under a unified Bayesian hyperparameter optimization protocol: GRU, LSTM, Bi-LSTM, 1D-CNN, Transformer, and a classification-adapted PatchTST. Each architecture was optimized independently using TPE (Tree-structured Parzen Estimator) with 100 search evaluations per seed across five seeds, yielding 3,000 total evaluations. The search space covered hidden size, dropout rate, learning rate, number of epochs, and batch size. The best configuration per seed was retrained and evaluated on the held-out test set.

Table 4. Architecture comparison with unified TPE optimization (PCA-Loading, k = 10, N = 7).

| Architecture | Macro F1 (%) | Parameters (mean) | Weight Size (KiB) | GPU Inference (ms/sample) |
|---|---|---|---|---|
| GRU | 98.29 +/- 0.62 | 138,905 | 545.90 | 0.38 |
| 1D-CNN | 98.08 +/- 1.28 | 76,201 | 300.17 | 0.29 |
| Bi-LSTM | 98.08 +/- 1.41 | 893,289 | 3,494.86 | 0.68 |
| Transformer | 97.00 +/- 1.87 | 754,333 | 2,955.14 | 1.03 |
| LSTM | 96.62 +/- 4.49 | 55,862 | 221.52 | 0.34 |
| PatchTST | 94.26 +/- 2.11 | 127,510 | 506.59 | 1.13 |

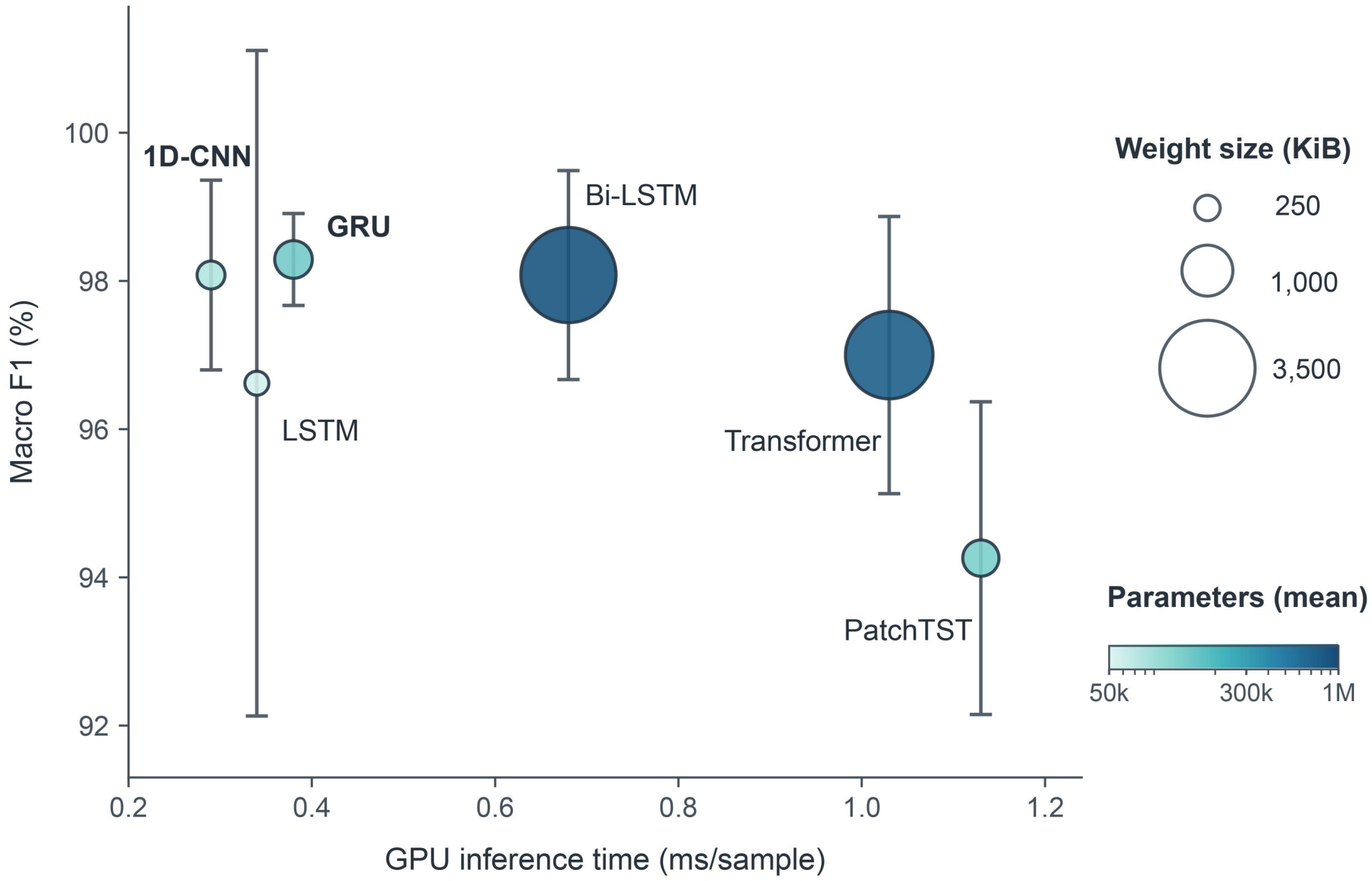

Fig. 8. Architecture comparison under unified TPE optimization. Vertical axis shows macro F1-score with error bars across five seeds. Horizontal axis shows GPU inference time. Bubble size indicates weight storage and color indicates parameter count.

Fig. 8 visualizes the trade-off between classification performance, inference speed, and model complexity. GRU occupies the upper-left region of the plot, combining the highest macro F1 (98.29 ± 0.62%) and the lowest standard deviation with moderate inference time and model size. The differences among GRU, 1D-CNN, and Bi-LSTM are within one standard deviation. 1D-CNN achieves comparable accuracy with the smallest parameter count and fastest inference but lacks explicit temporal modeling. Bi-LSTM is close to GRU in mean accuracy yet requires 6.4 times the weight storage and nearly twice the inference time. Transformer and PatchTST fall in the lower-right region, exhibiting both lower accuracy and higher computational cost. This pattern is consistent with the small-sample nature of this task. With only 734 training samples, self-attention may lack sufficient data to learn meaningful token interactions. GRU's sequential inductive bias provides a suitable structural prior for temporally ordered wear data. In this setting, recurrent models perform at least as well as attention-based models at lower cost.

To evaluate the effectiveness of hyperparameter optimization strategies, six algorithms were compared under equal evaluation budgets of 150 evaluations per seed across five seeds, yielding 4,500 total evaluations. All algorithms optimized the same GRU architecture with PCA-Loading k = 10 and N = 7.

Table 5. HPO algorithm comparison (GRU, PCA-Loading, k = 10, N = 7, 150 evaluations x 5 seeds).

| Method | Macro F1 (%) | Parameters (mean) | Search Time (s) |
|---|---|---|---|
| IPSO-SA | 98.38 +/- 1.39 | 206,517 | 214.76 |
| BO (TPE) | 98.21 +/- 1.52 | 195,703 | 311.17 |
| GA | 97.96 +/- 1.35 | 307,436 | 370.66 |
| Random Search | 97.80 +/- 1.44 | 238,096 | 224.39 |
| PSO | 96.58 +/- 3.65 | 383,248 | 264.06 |
| SA | 95.51 +/- 3.17 | 159,915 | 378.08 |

The TPE result in Table 5 (98.21% ± 1.52%) differs from Table 4 (98.29% ± 0.62%) because the two tables represent independently executed experiments; the difference falls within overlapping confidence intervals and reflects stochastic variation rather than a systematic discrepancy. IPSO-SA achieves the highest mean macro F1 of 98.38%. Its margin over standard PSO (96.58%) and standalone SA (95.51%) suggests a benefit from combining dynamic inertia weight adjustment with the SA-based acceptance mechanism. IPSO-SA performs on par with Bayesian optimization. The top four methods fall within 0.6 percentage points of each other, which is within one standard deviation. The fixed-hyperparameter GRU in Table 3 (98.47%) also lies in this range, so the classifier is not sensitive to hyperparameters in this task. Search time depends on the sampled epochs and hidden sizes and is reported for reference only.

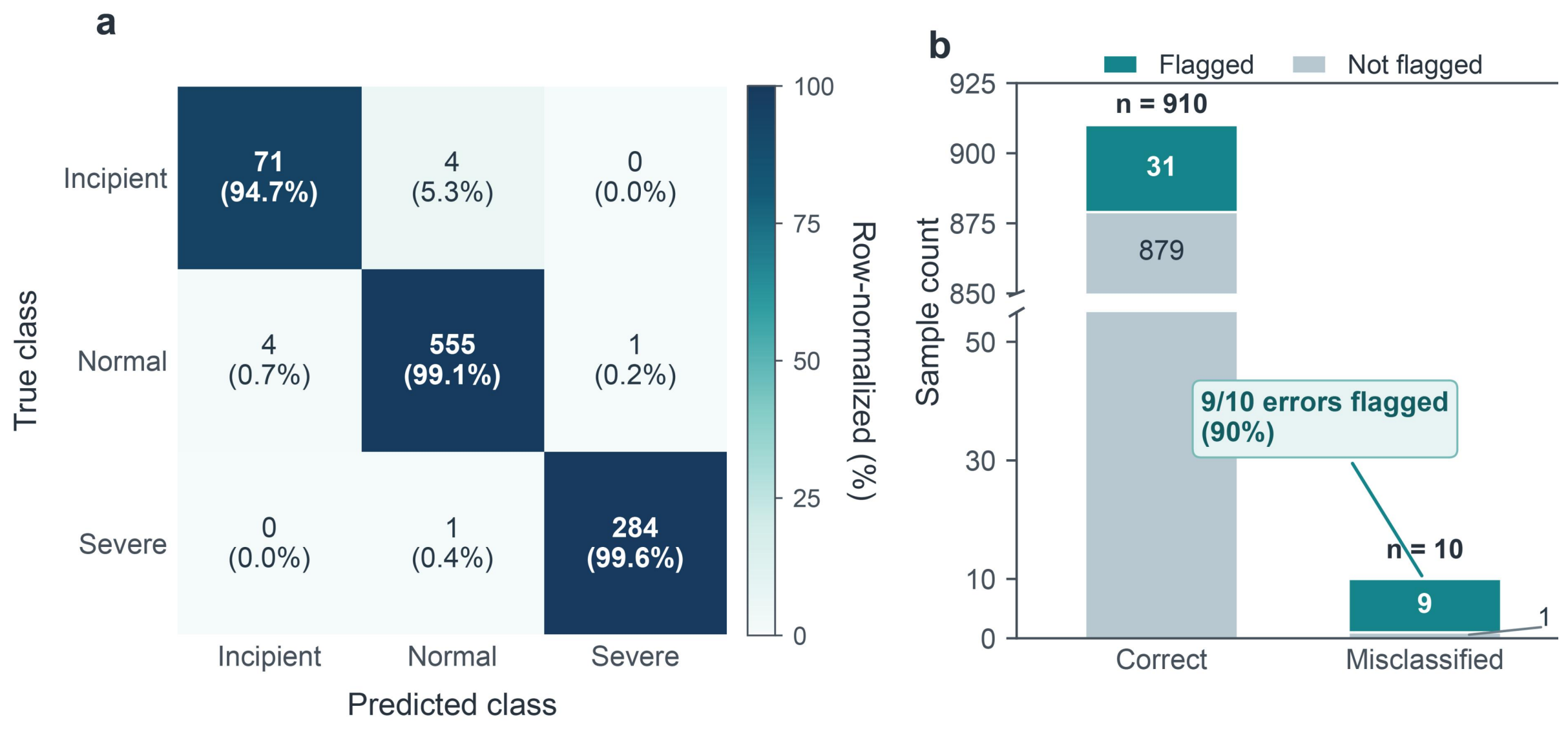

Fig. 9. IPSO-SA-GRU diagnostic performance aggregated across five test splits (920 predictions). (a) Normalized confusion matrix. (b) Review flag coverage.

Fig. 9(a) presents the aggregated confusion matrix of the IPSO-SA-GRU model. Misclassifications concentrate at the incipient-normal boundary, accounting for 8 of 10 total errors, with no cross-stage confusion between incipient and severe wear. This pattern is consistent with the gradual nature of wear progression, where adjacent states share similar feature distributions. Fig. 9(b) shows the review flag coverage: the mechanism identified 9 of the 10 misclassified predictions while flagging 31 correctly classified samples for precautionary verification. The single undetected error was a normal-to-severe misclassification with high softmax confidence. These results suggest that the review flag covers most errors of the diagnostic engine, although the number of errors is small.

To examine generalization to unseen tools, the final configuration was additionally evaluated under a leave-one-tool-out protocol (Table 6). Under the VB-based labels and triaxial vibration used in this study, the mean macro F1-score fell to 55.91%. The fixed VB thresholds correspond to different stages of tool life across tools: 0.08 mm is crossed at passes 16, 69, and 18, and 0.13 mm at passes 220, 250, and 191 for C1, C4, and C6, respectively. Adding force and acoustic emission channels did not improve this result (56.73%). When the pass-index stage boundaries of [50] were used instead, the same pipeline with all seven channels reached a mean macro F1-score of 92.00%, comparable to values reported under this protocol [50], whereas vibration alone reached 50.37%, mainly because the late-stage vibration amplitude of C6 was markedly higher than that of the training tools (mean X-axis RMS over passes 176–315 of 0.222 for C6 versus 0.132 and 0.133 for C1 and C4). The results in Tables 3–5 therefore reflect recognition within the wear trajectories of the three tools, and cross-tool generalization with VB-based labels and vibration-only input remains limited. Because the reasoning layer operates on the evidence record regardless of classifier accuracy, this limitation does not affect the evaluation in Section 4.4.

Table 6. Leave-one-tool-out macro F1-score (%) under different label definitions and input channels (N = 7).

| Label definition | Input channels | Test C1 | Test C4 | Test C6 | Mean |
|---|---|---|---|---|---|
| VB-based (0.08/0.13 mm, this study) | Triaxial vibration | 47.42 | 64.74 | 55.57 | 55.91 |
| VB-based (0.08/0.13 mm) | All seven channels | 68.08 | 64.49 | 37.61 | 56.73 |
| Pass-index stages | Triaxial vibration | 72.86 | 53.87 | 24.39 | 50.37 |
| Pass-index stages | All seven channels | 91.02 | 97.80 | 87.18 | 92.00 |

Note:Mean over five random seeds. In each fold, two tools were used for training and the third for testing; PCA-Loading feature selection (top 10) and z-score normalization were fitted on the training tools only. Pass-index stages follow [50]: passes 1–50, 51–175, and 176–315. The first row corresponds to the configuration used in this study.

## 4.4 LLM-based reasoning evaluation

The reasoning layer was evaluated on 552 PHM2010 evidence records generated by the k = 10 IPSO-SA-GRU model from seeds 44, 45, and 46 (184 test samples per seed). For each record, the ablation procedure was applied to the 10 input features across 7 time steps, temporal trends were computed for the top 5 features, and the review flag was determined by GRU-RF comparison. All records were processed under four conditions using DeepSeek V4 Flash (temperature 0, max tokens 2048), each record processed once with no retries.

Conditions A to C form an ablation in which evidence and prompt constraints are added step by step. Condition A provides only the predicted wear state and softmax probability, with a single-sentence prompt. Condition B provides the complete evidence record with a minimal prompt requesting a diagnostic output. Condition C is the full pipeline with designed prompt and pre-computed factual constraints. Conditions D and E are the basic and enhanced rule templates (Section 3.3.2). The results are summarized in Table 7.

Table 7. Factual accuracy comparison across prompt conditions (DeepSeek V4 Flash, 552 PHM2010 outputs).

| Condition | Evidence provided | Prompt design | Numerical fidelity | Factual consistency | Completeness | Fully faithful |
|---|---|---|---|---|---|---|
| A | Prediction + confidence only | Minimal | 345 / 552 | — | — | — |
| B | Full evidence record | Minimal | 126 / 552 | 126 / 552 | 126 / 552 | 0 / 552 |
| C | Full evidence record | Designed | 550 / 552 | 550 / 552 | 550 / 552 | 550 / 552 |
| D | Full evidence record | Rule template | 552 / 552 | 552 / 552 | 552 / 552 | 552 / 552 |

Condition A demonstrates that minimal diagnostic input leads to substantial deficiencies. Of 552 outputs, 207 did not meet the numerical fidelity requirement, including 203 non-parseable outputs, as the minimal prompt did not specify an output format. Among the 349 parseable outputs, 228 contained unsupported feature-attribution or trend claims with no basis in the provided input. Factual consistency, completeness, and full faithfulness are undefined for Condition A because no feature-level evidence was provided for verification.

Condition B shows that providing the complete evidence record alone is insufficient. The lower numerical fidelity compared to Condition A (126 versus 345) reflects the compounding effect of a larger verification surface: approximately 20 checkable numerical fields versus 2, where a single error fails the entire output. No output passed all checks: the same 126 outputs passed the first three checks, but none reproduced the supplied action code, as the minimal prompt did not require it.

Condition C passes all checks for 550 of 552 outputs; the two failures were non-parseable outputs, and all 550 parseable outputs were fully faithful. No unsupported claims or transcription errors were observed under the designed prompt with pre-computed factual constraints.

To assess whether this result depends on the specific LLM, the same 552 records were processed under Condition C using Claude Sonnet 4.6 and Qwen3.8-Max (Table 8). The pipeline was also applied to 146 NASA Ames records generated through stratified 5-fold out-of-fold evaluation with Random Forest classifiers on 168-dimensional features (6 channels times 28 features).

Table 8. Factual accuracy across LLMs under the full pipeline (Condition C, 552 PHM2010 outputs).

| **LLM** | **API timeouts** | **Non-parseable outputs** | **Numerical fidelity** | **Factual consistency** | **Completeness** | **Fully faithful** |
|---|---|---|---|---|---|---|
| DeepSeek V4 Flash | 0 | 2 | 550 / 552 | 550 / 552 | 550 / 552 | 550 / 552 |
| Claude Sonnet 4.6 | 7 | 1 | 543 / 552 | 544 / 552 | 544 / 552 | 543 / 552 |
| Qwen3.8-Max | 1 | 0 | 551 / 552 | 551 / 552 | 551 / 552 | 551 / 552 |
| **Cross-dataset** | | | | | | |
| DeepSeek (NASA, RF, n = 146) | 0 | 0 | 146 / 146 | 146 / 146 | 146 / 146 | 146 / 146 |

All three LLMs achieve above 98% factual accuracy. Each record was submitted once, with no transport or content retries. API timeouts and non-parseable outputs were retained and counted as failures in the fixed denominator. Of the twelve failures, eleven were API timeouts or non-parseable outputs; the only content error was one numerical inconsistency in a Claude Sonnet 4.6 report. The NASA result shows that the reasoning layer preserves factual accuracy across substantially different configurations: 250 Hz mixed six-channel signals versus 50 kHz triaxial vibration, 168 versus 10 input features, RF versus GRU, and 83% versus 98% macro F1-score. The review flag trigger rate on NASA (30.8%, 45 of 146) contrasts with PHM2010 (4.5%), and the flag captured 24 of the 25 misclassified NASA records, including all nine severe-wear records predicted as a lower wear state (Fig. 10). The reasoning layer reliably communicates this elevated uncertainty.

Tables 7 and 8 show that the full pipeline achieves factual accuracy comparable to a deterministic template. To evaluate whether the LLM provides additional value beyond factual correctness, a blind evaluation compared Condition C against Conditions D and E. One hundred records were sampled from the 552, stratified by wear state (30 incipient, 30 normal, and 40 severe). Three CNC engineers, each with over ten years of industry experience, independently rated each report on a five-point scale (1 = very poor, 5 = very good) for actionability (whether the report indicates what to do next), overall quality, clarity, and decision confidence (how confident the evaluator would be in making a replacement decision based on the report). The three reports of each pair were presented in randomized order without method labels, and the true wear state was not disclosed. The score of each report was the mean of the three ratings, and differences of C from D and from E over the 100 records were tested with the Wilcoxon signed-rank test with Holm correction.

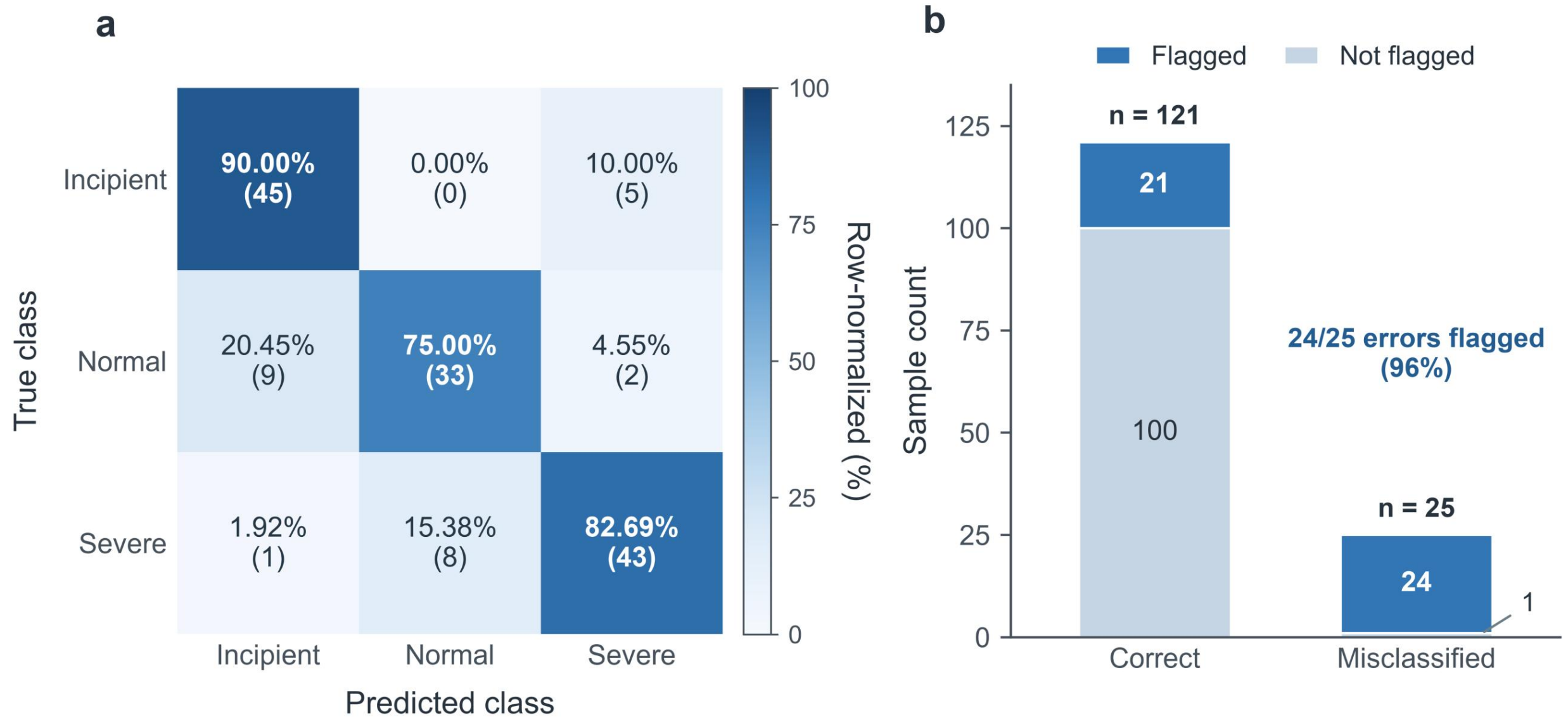

Fig. 10. RF diagnostic performance on NASA Ames aggregated over five out-of-fold splits (146 predictions). (a) Row-normalized confusion matrix. (b) Review flag coverage.

Table 9 shows that LLM reports are rated higher than both templates across all four metrics, with no 95% confidence interval including zero (C versus E, Holm-adjusted $p < 0.001$). Decision confidence receives the lowest basic-template score (2.98), indicating that listing feature values independently does little to help an engineer decide whether to replace the tool. The enhanced template raises this score to 3.58 and the LLM report to 3.92. Across the four metrics, E recovers 59–77% of the improvement of C over D, so grouping features by effect, summarizing trends, and matching urgency to the action code account for most of the benefit. The remaining gap is smallest for actionability (+0.21), where both conditions follow the same pre-computed action code, and largest for overall quality (+0.43).

Table 9. Human evaluation of LLM reports versus basic and enhanced rule templates (100 records, 3 evaluators; differences are paired mean differences with 95% CIs).

| Metric | Rule template (D), mean ± SD | Rule template (E), mean ± SD | LLM report (C), mean ± SD | Δ (C minus D) [95% CI] | Δ (C minus E) [95% CI] | Holm-adjusted p | Rank-biserial r | Ordinal α, D / C |
|---|---|---|---|---|---|---|---|---|
| Actionability (1-5) | 3.12 ± 0.51 | 3.82 ± 0.47 | 4.03 ± 0.43 | +0.91 [0.79, 1.03] | +0.21 [0.09, 0.33] | < 0.001 | 0.82 | 0.78 / 0.85 |
| Overall quality (1-5) | 3.06 ± 0.48 | 3.67 ± 0.55 | 4.10 ± 0.40 | +1.04 [0.92, 1.16] | +0.43 [0.30, 0.56] | < 0.001 | 0.86 | 0.80 / 0.87 |
| Clarity (1-5) | 3.18 ± 0.50 | 3.77 ± 0.49 | 4.16 ± 0.39 | +0.98 [0.84, 1.12] | +0.39 [0.26, 0.52] | < 0.001 | 0.84 | 0.81 / 0.88 |
| Decision confidence (1-5) | 2.98 ± 0.56 | 3.58 ± 0.52 | 3.92 ± 0.46 | +0.94 [0.81, 1.07] | +0.34 [0.20, 0.48] | < 0.001 | 0.79 | 0.76 / 0.84 |

Three misclassifications were encountered during the evaluation. In each case the review flag correctly identified the prediction as requiring verification: two were flagged due to GRU-RF disagreement and one due to low softmax confidence of 0.62. The LLM outputs conveyed the flag status in all three cases without attempting to override the classification. LLM-based reasoning averages 10.03 seconds per output (P95: 11.19 s) versus less than 0.01 seconds for the template, which is acceptable given inter-pass intervals of 30 seconds to several minutes.

## 4.5 Case study

To demonstrate how the proposed framework supports tool replacement decisions in practice, three representative samples are selected and summarized in Table 10. The samples cover three scenarios that a CNC operator would encounter during production: a confident severe-wear alert (Case 1), a borderline incipient-wear detection where the system requests operator review before action (Case 2), and a misclassification on a different machining setup where the safety mechanism correctly flags the prediction as uncertain (Case 3). Case 2 is presented in detail in Fig. 11.

Fig. 11 shows the complete pipeline for Case 2. After the 18th cutting pass of tool C4, the diagnostic engine classifies the wear state as incipient with a GRU confidence of only 0.5495.

Although the independent Random Forest classifier agrees on the incipient state with higher confidence (0.88), the low GRU confidence triggers the review flag. The evidence record reveals conflicting vibration signatures: the most influential feature, z-axis mean vibration (vib_z_mean, ablation delta of minus 3.71), actively weakens the incipient classification, while x-axis mean vibration (vib_x_mean, plus 3.52) supports it. Overall, three of the five top features work against the prediction. Furthermore, four of the five features show negative slopes over the 7-pass monitoring window despite positive first-to-last changes, indicating that these features rose early in the window and then declined, so no sustained upward trend is present. For an operator at the machine, this mixed pattern makes an immediate tool change unnecessary but warrants closer inspection.

The LLM report generated under Condition C communicates this situation in a form suitable for operator decision-making. Like the enhanced template, it groups the five features by their effect on the classification, noting that two support and three weaken the incipient prediction, and it further relates this conflict to the reversing trends. It preserves the caveat that the classification is a model output rather than a direct flank wear measurement, and it produces the REVIEW_NOW action code. In practice, this would prompt the operator to inspect the tool before the next cutting pass rather than replacing it immediately. All numerical values in the output match the evidence record exactly, consistent with Table 7, in which all 550 parseable outputs under Condition C were fully faithful.

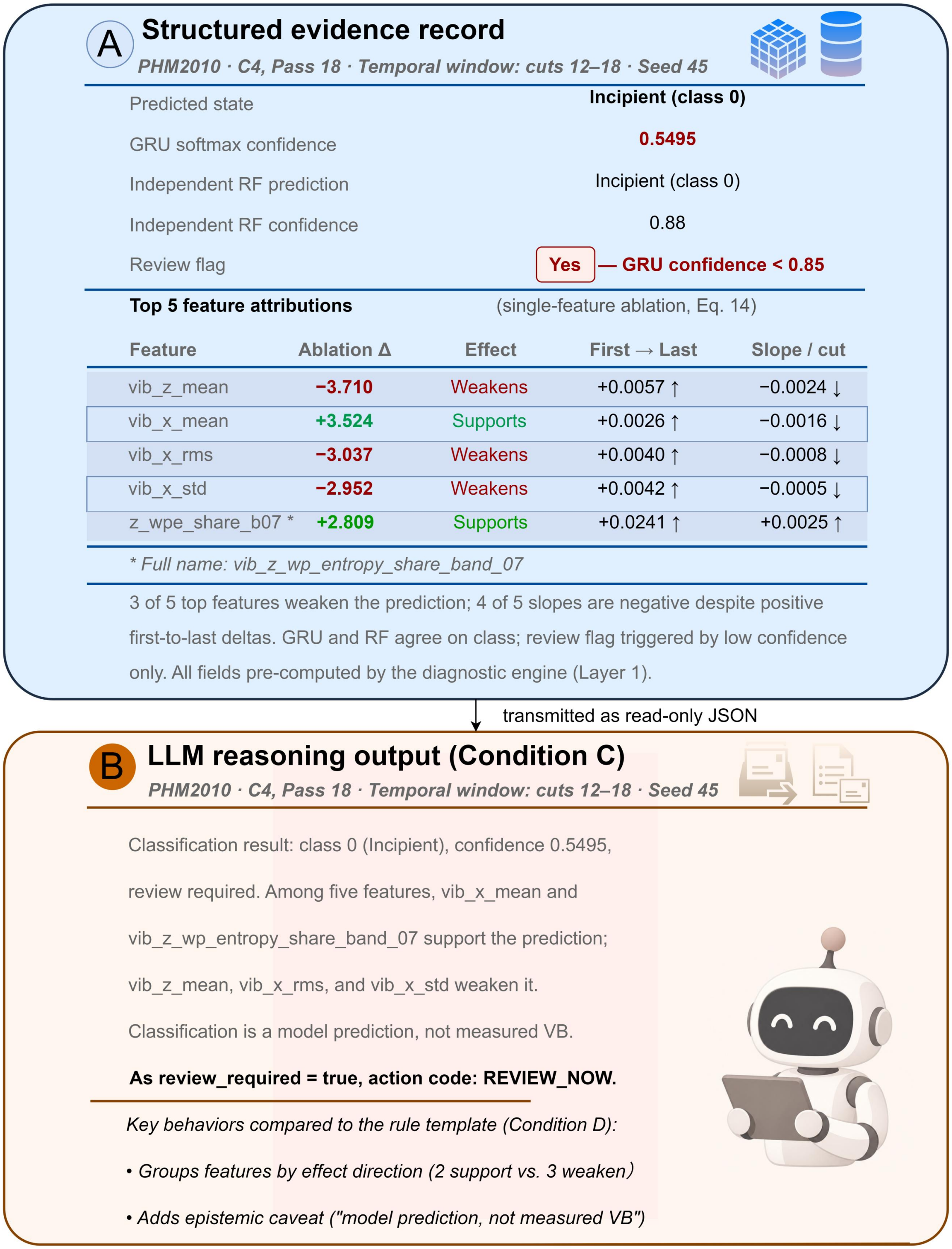

Fig. 11. Evidence record and LLM reasoning output for a review-flagged incipient wear sample (PHM2010, C4, Pass 18).

Table 10. Case study summary across three diagnostic scenarios.

| Item | Case 1 | Case 2 (Fig. 11) | Case 3 |
|---|---|---|---|
| Dataset | PHM2010 | PHM2010 | NASA Ames |
| Sample | --- | C4, Pass 18 | --- |
| True state (measured VB) | Severe (0.1306 mm) | Incipient (0.0629 mm) | Normal (0.2800 mm) |
| Predicted state | Severe ✓ | Incipient ✓ | Severe ✗ |
| Softmax / RF confidence | 0.95 | 0.55 | 0.68 |
| Review flag | No | Yes | Yes |
| Flag trigger | --- | GRU confidence < 0.85 (GRU and RF agree) | RF confidence < 0.85 (single-classifier) |
| Key evidence pattern | Mixed support and weaken attributions | 3 of 5 features weaken; conflicting trends | All top features support incorrect prediction |
| Rule template (Cond. D) | Five features listed independently; uniform tone | (same) | (same) |
| LLM report (Cond. C) | Synthesizes evidence; recommends tool replacement | Groups features by effect; recommends VB measurement | Reports prediction with flag; recommends manual verification |

Note. Case 3 uses a different diagnostic architecture (168-dim RF, no GRU or temporal sequences) and illustrates cross-dataset transferability.

The three cases in Table 10 show that the reasoning layer adapts its communication to match the diagnostic situation. In Case 1, the model predicts severe wear with 0.95 confidence and no review flag. The LLM report conveys this with appropriate urgency and recommends tool replacement before the next cutting pass. Because the severe grade is defined relative to the VB range of the dataset rather than an ISO tool-life criterion (Section 4.1), the replacement threshold would be set according to shop-specific tool-life criteria in deployment. In Case 3, the model incorrectly predicts severe wear for a sample whose actual flank wear (0.2800 mm) falls within the normal range under the NASA Ames thresholds. The review flag correctly identifies this prediction as uncertain. The LLM does not attempt to override the classification; it faithfully communicates the severe prediction alongside the review flag and recommends manual verification, leaving the replacement decision to the operator. This is the intended behavior: the diagnostic engine determines all factual content, and the reasoning layer is responsible for organizing and communicating the evidence rather than making independent diagnostic judgments.

Case 3 also illustrates cross-dataset transferability. The NASA Ames diagnostic engine uses a 168-dimensional Random Forest on single-pass features without temporal sequences, which is a fundamentally different architecture from the PHM2010 GRU pipeline. The reasoning layer processed this different evidence structure without any modification, and all 146 NASA Ames reports achieved complete factual accuracy. This result suggests that the structured evidence interface can accommodate different machining setups, sensor configurations, and classifier architectures without requiring changes to the reasoning layer.

## 4.6 Discussion

The Pareto-optimal subset of 10 features identified by PCA-Loading retains 99.66% of the peak classification performance achieved with all 84 features. Examination of the selected features (Fig. 5) reveals that 8 of 10 are time-domain amplitude statistics (the mean and standard deviation of all three axes and the RMS of the x and y axes), with the remaining two drawn from the frequency domain (z-axis spectral RMS frequency) and the time-frequency domain (z-axis wavelet packet entropy in sub-band 7). The dominance of time-domain descriptors indicates that, at this level of feature compression, the classifier relies mainly on basic amplitude statistics rather than spectral or time-frequency representations. Physically, RMS reflects the overall vibration energy level, which is expected to increase as the flank wear land grows and cutting forces rise. Standard deviation captures the variability of the vibration waveform. The mean of an accelerometer signal mainly reflects sensor offset and slow drift, so no physical meaning is assigned to it here. Because such drift accumulates with cutting time, the mean features may partly act as a proxy for elapsed cutting time within a tool trajectory, which may contribute to the gap between the within-trajectory and leave-one-tool-out results. Because the reasoning layer is not permitted to introduce physical causal mechanisms (Appendix A), these features are reported to the operator as model evidence rather than as indicators of wear physics. The three axes contribute relatively evenly (3 from x, 3 from y, 4 from z). However, the ablation attribution analysis for Case 2 (C4, Pass 18) reveals that the x and z axes dominate the top five most influential features, while no y-axis feature appears among them, suggesting that the y-axis features contribute to overall classification performance but may play a less decisive role in individual boundary-case predictions. Among the top five attributions, vib_z_mean produces the largest ablation delta (−3.71) but weakens the incipient classification, while vib_x_mean (+3.52) supports it. Single-feature ablation does not account for correlation among inputs. Correlated features can therefore receive deltas of opposite sign without opposite physical effects. The attributions describe model behavior and are not evidence of the wear mechanism. The presence of vib_z_wp_entropy_share_band_07 (+2.81) among the supporting features suggests that the z-axis wavelet sub-band adds information beyond time-domain statistics. PCA-Loading is unsupervised, so its advantage over RF Importance and Mutual Information at low feature counts is an empirical observation for this dataset. The selected subset also contains strongly correlated pairs, such as standard deviation and RMS of the same axis, so its effective dimension is lower than ten.

The aggregated confusion matrix in Fig. 9(a) reveals that 8 of 10 misclassifications occur at the incipient-normal boundary, with no cross-stage confusion between incipient and severe wear. This error distribution is consistent with the physical characteristics of progressive flank wear illustrated in Fig. 2. The transition from incipient to normal wear corresponds to the inflection region between rapid initial wear and the prolonged steady-state growth phase, where the VB change gradient is relatively small and the vibration signal characteristics of adjacent cutting passes may be similar. By contrast, the transition from normal to severe wear coincides with accelerating degradation, producing more pronounced signal changes that are easier to distinguish. Incipient wear is also the minority class (10.6% of samples), so its recall is the metric most sensitive to class imbalance. The selection of GRU over Transformer and PatchTST architectures can also be understood from this perspective. Tool wear is an inherently cumulative process where flank wear grows monotonically and the current wear state depends on the entire preceding cutting history. GRU's sequential inductive bias provides a structural prior that naturally aligns with this monotonic temporal progression, whereas self-attention mechanisms lack such prior assumptions and may require substantially more training samples to learn meaningful temporal dependencies. With only 734 training samples in this study, the recurrent architecture's built-in temporal bias appears to compensate for limited data availability. This advantage was observed within the tool trajectories; under leave-one-tool-out evaluation, a single-pass RF performed comparably (macro F1-score 58.79% versus 55.91%), so the benefit of temporal context did not extend to unseen tools. The hyperparameter optimization results can be understood from the same small-sample perspective. Industrial tool wear datasets typically contain fewer than one thousand labeled samples per tool due to the cost of optical wear measurement, which limits the validation set size and amplifies evaluation variance. In this regime, differences among search strategies are small relative to this variance: the top four methods in Table 5 and the fixed-hyperparameter GRU in Table 3 lie within one standard deviation of each other. IPSO-SA showed smaller variance than standard PSO and standalone SA, but the main implication for the framework is that the evidence passed to the reasoning layer does not depend on a particular tuning strategy.

In safety-critical manufacturing environments, a diagnostic report containing fabricated or inaccurate content could lead to missed tool replacements that risk workpiece scrap or spindle damage, or to unnecessary production interruptions that reduce machine utilization. The controlled evaluation across Conditions A through D provides empirical justification for the design choices made to address this concern. Under Condition A, where the LLM received only the predicted wear state and confidence, 207 of 552 reports did not meet the numerical fidelity requirement, and 228 of the 349 parseable reports fabricated feature attribution claims with no basis in the provided input, demonstrating that LLMs can generate plausible but unsupported technical content when diagnostic evidence is insufficient. Condition B supplied the complete evidence record but used only a minimal prompt, and its numerical fidelity dropped below that of Condition A due to the expanded verification surface. In this study, complete faithfulness was observed only when both the complete evidence record and the designed prompt with pre-computed factual fields were present (Condition C). Single-feature ablation follows the same reliability principle. It gives identical output for identical input, so every evidence field is exactly reproducible. This deterministic property is particularly important when the evidence record serves as a factual constraint for LLM generation, as any variability in the input could propagate into inconsistent natural language outputs. The consistency of this design across three LLMs further suggests that the reliability depends primarily on the structured interface rather than on any specific model's capabilities, which may facilitate adaptation to future model versions without re-engineering the evidence pipeline.

The cross-dataset validation between PHM2010 and NASA Ames provides preliminary evidence that the reasoning layer can accommodate substantially different machining configurations. The two datasets differ in workpiece material, spindle speed (10,400 rpm versus 826 rpm), sensor type and sampling rate (50 kHz triaxial vibration versus 250 Hz six-channel signals including acoustic emission and spindle current), and diagnostic architecture (10-feature GRU with temporal sequences versus 168-feature Random Forest on single-pass vectors). Despite these differences, all 146 NASA Ames reports achieved complete factual accuracy under the same prompt template without modification. The review flag trigger rate rose from 4.5% on PHM2010 to 30.8% on NASA Ames, reflecting the lower diagnostic performance of the NASA configuration (macro F1-score 83% versus 98%) while still capturing nearly all misclassified records. This behavior suggests that the review flag mechanism functions as intended: it does not assume high diagnostic accuracy but faithfully communicates the level of certainty present in the evidence record. The human evaluation results in Table 9 further illuminate why structured natural language reports may offer practical value beyond deterministic rule templates. Evaluators rated LLM reports higher than both templates although all three contain the same facts. The enhanced template, which groups features by effect, summarizes trends, and adjusts urgency with deterministic rules, recovered most of this improvement. Information organization, rather than information content, is therefore the primary driver, and much of it does not require an LLM. The remaining advantage of the LLM likely lies in relating attributions and trends to each other within a single interpretation, which dimension-wise rules do not cover. Because all facts and actions are fixed before generation, this design secures reliability but also bounds the value the LLM can add, suggesting that how diagnostic evidence is communicated may be as important as the evidence itself for supporting practical replacement decisions.

# 5. Conclusions

This study addressed the challenge of generating interpretable and reliable diagnostic reports for tool wear monitoring in CNC machining, where deep learning classifiers achieve high accuracy but produce predictions that are hard for operators to act upon directly.

A two-layer architecture was proposed that decouples diagnostic computation from natural language generation through a structured evidence record interface. The first layer, the evidence-augmented diagnostic engine, processes triaxial vibration signals through multi-domain feature extraction, PCA-Loading feature selection, and GRU-based temporal classification with IPSO-SA hyperparameter optimization. A systematic evaluation comprising 540 feature selection runs, 3,000 architecture comparison runs, and 4,500 hyperparameter optimization runs on the PHM2010 dataset identified the Pareto-optimal configuration of k = 10 features and N = 7 sequence length, achieving 98.38% macro F1-score. The selected 10-feature subset retains 99.66% of the peak performance obtained with all 84 features, suggesting that the dominant discriminative information is concentrated in a small number of original features. Six hyperparameter optimization methods, including the IPSO-SA variant used in the pipeline, produced similar results, indicating that the classifier is not sensitive to hyperparameter settings in this task. Leave-one-tool-out evaluation further showed that the fixed VB thresholds correspond to different stages of tool life across tools, limiting cross-tool generalization with vibration-only input (mean macro F1-score 55.91%), whereas the same pipeline reached 92.00% with all seven channels under pass-index stage labels.

The second layer, the LLM-based multi-evidence reasoning layer, receives pre-computed classification results, feature attributions, temporal trends, and cross-model review flags as pre-computed factual fields, confining LLM generation to narrative coherence, contextual interpretation, and action suggestions. Evaluation across 552 PHM2010 reports under controlled conditions showed no factual errors in the 550 parseable reports, whereas 228 of 349 parseable reports generated without structured evidence contained unsupported feature-attribution or trend claims. Blind evaluation by three experienced CNC engineers rated LLM reports above both a basic and an enhanced rule template, with the enhanced template recovering 59–77% of the improvement, indicating that evidence organization drives most of the benefit while the LLM adds a further gain. Cross-dataset validation on 146 NASA Ames records with a fundamentally different signal modality, classifier architecture, and accuracy level indicated that the reasoning layer can be applied without modification. The review flag trigger rate rose from 4.5% to 30.8%, and all reports faithfully communicated this elevated uncertainty.

Several limitations remain. First, the diagnostic engine was validated on the PHM2010 and NASA Ames public datasets rather than through field deployment in operational CNC workshops. Future work will focus on long-term validation under real-world conditions including noise interference and workpiece variability, on domain adaptation to improve cross-tool generalization of the diagnostic engine, and on excluding offset-related features such as the accelerometer mean, which may act as proxies for cutting time. Second, the LLM evaluation used three models, deterministic decoding, and a fixed prompt template. Robustness under other models, sampling temperatures, and prompt variations remains to be tested, and allowing labeled, traceable inferences beyond the pre-computed evidence may further increase the value of the reasoning layer. Third, the review flag mechanism relies on a fixed confidence threshold and a single secondary classifier. Future work will explore adaptive threshold calibration based on accumulated operational data and ensemble-based review strategies to improve detection of uncertain predictions across diverse machining conditions.

## Data availability statement

The data used in this study is available from the PHM Society 2010 Data Challenge (https://phmsociety.org/phm10-data-challenge-leaderboard/). The NASA Ames Milling dataset is available from the NASA Open Data Portal (https://data.nasa.gov/dataset/milling-wear).

## Declaration of competing interest

The authors declare that they have no known competing financial interests or personal relationships that could have appeared to influence the work reported in this paper.

## CRediT authorship contribution statement

Yuntong Chen: Writing – original draft, Software, Methodology, Formal analysis, Data curation, Visualization, Conceptualization. Yingqi Li: Writing – review & editing, Validation. Qiwen Shen: Software, Validation, Data curation. Guobin Zhao: Formal analysis, Validation. Xitian Tian: Supervision, Formal analysis. Lijiang Huang: Writing – review & editing, Supervision, Funding acquisition, Conceptualization.

## Declaration of generative AI and AI-assisted technologies in the manuscript preparation process

During the preparation of this work the authors used Chat GPT 5.6 Sol (OpenAI) in order to improve the readability and language quality of the manuscript. After using this tool, the authors reviewed and edited the content as needed and take full responsibility for the content of the published article.